\documentclass[11pt,a4paper]{article}
\usepackage{times,latexsym}
\usepackage{float}

\usepackage{url}
\usepackage[T1]{fontenc}

\usepackage[acceptedWithA]{tacl2021v1}

\usepackage{xspace,mfirstuc,tabulary}
\newcommand{\dateOfLastUpdate}{Dec. 15, 2021}
\newcommand{\styleFileVersion}{tacl2021v1}

\newif\iftaclinstructions
\taclinstructionsfalse %
\iftaclinstructions
\renewcommand{\confidential}{}
\renewcommand{\anonsubtext}{(No author info supplied here, for consistency with
TACL-submission anonymization requirements)}
\newcommand{\instr}
\fi

\iftaclpubformat %

\newcommand{\TaclPapers}{Final Versions\xspace}
\else

\newcommand{\TaclPapers}{Submissions\xspace}
\fi

\newcommand{\rulesep}{\unskip\ \vrule\ }

\usepackage{amsmath,amsfonts,bm}

\def\eqref#1{equation~\ref{#1}}

\def\1{\bm{1}}

\DeclareMathAlphabet{\mathsfit}{\encodingdefault}{\sfdefault}{m}{sl}
\SetMathAlphabet{\mathsfit}{bold}{\encodingdefault}{\sfdefault}{bx}{n}

\usepackage{graphicx}
\usepackage{comment}
\usepackage{subcaption}
\usepackage{algorithm}
\usepackage{algorithmic}
\usepackage{amsmath}
\usepackage{verbatim}
\usepackage{tabularx}
\usepackage[usestackEOL]{stackengine}
\strutlongstacks{T}

\definecolor{positive}{HTML}{F1A340}
\definecolor{negative}{HTML}{998EC3}

\title{Formatting Instructions for TACL \TaclPapers \\
(Base files: \styleFileVersion-template.tex \& \styleFileVersion.sty, dated \dateOfLastUpdate)}

\newcommand{\neu}{1}
\newcommand{\kcl}{2}

\author{
Jay DeYoung$^{\neu}$~\;~Stephanie C. Martinez$^{\neu}$~\;~Iain J. Marshall$^{\kcl}$~\;~Byron C. Wallace$^{\neu}$ \\
$^{\neu}$Northeastern University, Boston, MA~\;~$^{\kcl}$King's College London, London \\
\texttt{\href{mailto:deyoung.j@northeastern.edu}{deyoung.j@northeastern.edu}}~\;~\texttt{\href{mailto:martinez.s@northeastern.edu}{martinez.s@northeastern.edu}} \\
\texttt{\href{mailto:iain.marshall@kcl.ac.uk}{iain.marshall@kcl.ac.uk}}~\;~\texttt{\href{b.wallace@northeastern.edu}{b.wallace@northeastern.edu}}
}

\title{Do Multi-Document Summarization Models \emph{Synthesize}?}

\begin{document}
\maketitle
\begin{abstract}

Multi-document summarization entails producing concise synopses of collections of inputs. 
For some applications, the synopsis should accurately \emph{synthesize} inputs with respect to a key aspect, e.g., a synopsis of film reviews written about a particular movie should reflect the average critic consensus.
As a more consequential example, narrative summaries that accompany biomedical \emph{systematic reviews} of clinical trial results should accurately summarize the potentially conflicting results from individual trials.
In this paper we ask: To what extent do modern multi-document summarization models implicitly perform this sort of synthesis?
We run experiments over opinion and evidence synthesis datasets using a suite of summarization models, from fine-tuned transformers to GPT-4.
We find that existing models partially perform synthesis, but imperfectly: even the best performing models are over-sensitive to changes in input ordering and under-sensitive to changes in input compositions (e.g., ratio of positive to negative reviews). 
We propose a simple, general, effective method for improving model synthesis capabilities by generating an explicitly diverse set of candidate outputs, and then selecting from these the string best aligned with the expected aggregate measure for the inputs, or \emph{abstaining} when the model produces no good candidate.
\end{abstract}

\section{Introduction} \label{sec:intro}

\begin{figure*}\centering
\includegraphics[scale=0.275]{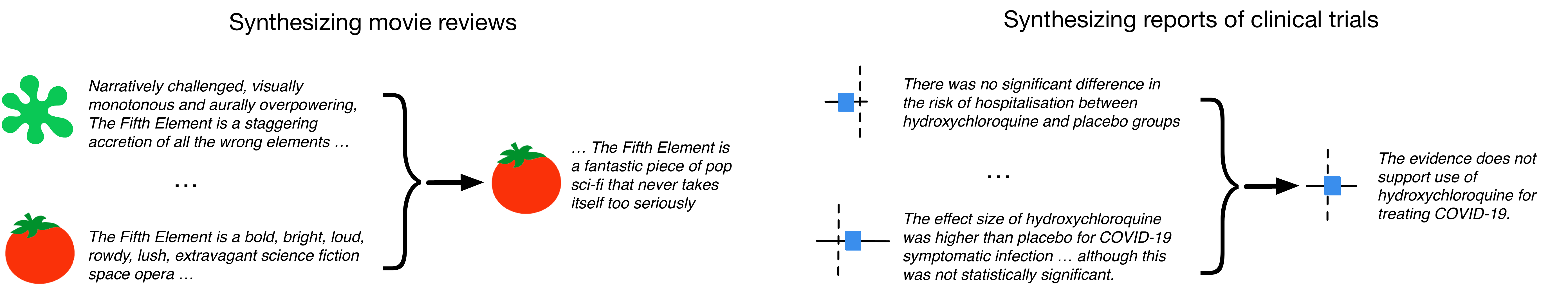}
    \caption{Two multi-document summarization tasks where models must implicitly synthesize inputs to produce accurate summaries. Left: Summarizing film reviews with varying sentiment to yield a \emph{critics consensus}. Right: Summarizing trials that have evaluated a particular medical invention.}
    \label{figure:synthesis_as_summarization}
\end{figure*}

\begin{table*}[t]
\small
    \centering
    \begin{tabular}{p{0.7\linewidth} p{0.25\linewidth}}
    Study &  Predicted Effect \\\hline
        \textbf{Input:} ...Ibuprofen was twice as
likely as acetaminophen to abort migraine within 2 hours. In the intent-to-treat 
analysis, children improved twice as often with ibuprofen and acetaminophen as
with placebo... & no significant difference \\
       \textbf{Input:} ...Children's ibuprofen suspension at an OTC dose of 7.5 mg/kg is an
effective and well-tolerated agent for pain relief in the acute treatment of
childhood migraine, particularly in boys... & significant difference\\\hline
\textbf{Target:} ...Low quality evidence from two small trials shows that ibuprofen appears to improve pain freedom for the acute treatment of children with migraine. We have only limited information on adverse events associated with ibuprofen in the trials included in this review... & no significant difference \\\hline
    \end{tabular} %
    \caption{Systematic review example (from Cochrane). The  statistical meta-analysis result "significant difference" and {\tt RobotReviewer} finding "no significant difference" disagree. In the case of Systematic Reviews, {\tt RobotReviewer} serves as both the estimator of $z_{ij}$ and $G$.}
    \label{table:systematic_review_example}
    \vspace{-0.5em}
\end{table*}

\emph{Multi-document summarization} (MDS) models aim to distill inputs into concise synopses that preserve key content. 
Examples of MDS include summarizing news articles \citep{Dang2005OverviewOD,fabbri-etal-2019-multi,gholipour-ghalandari-etal-2020-large,evans-etal-2004-columbia}, answering questions from multiple sources \citep{Dang06overviewof}, and producing overviews of scientific literature  \citep{Liu2018GeneratingWB,lu-etal-2020-multi-xscience,molla2012creation,Wallace2020GeneratingN,deyoung-etal-2021-ms}.  
We expect summarization models to produce outputs consistent with inputs \citep{kryscinski-etal-2020-evaluating,nan-etal-2021-improving}, e.g., discussing the same types of entities \citep{nan-etal-2021-entity} and allowing one to answer questions similar in a way that is consistent with individual inputs \citep{wang-etal-2020-asking,scialom-etal-2021-questeval}.

In some applications models must \emph{synthesize} inputs---i.e., aggregate potentially conflicting information---to yield an accurate synopsis (Figure \ref{figure:synthesis_as_summarization}).
Consider the meta-reviews of movies featured on Rotten Tomatoes,\footnote{\url{https://www.rottentomatoes.com/}.} which provide a consensus view of individual critic opinions. 
These reviews should reflect the mean and range of sentiment implicit in the input critiques: A summary of mostly negative reviews (e.g., \emph{Gigli}) should communicate that the film was widely panned; a summary of mixed reviews (\emph{The Fifth Element}) ought to convey that critics disagreed and discuss the main positive and negative attributes.

A more consequential example is summarizing the evidence presented in clinical trials.
Individual trials will often present conflicting evidence about whether or not a particular health intervention is effective.
An ideal summary of %
would appropriately weigh the findings presented in individual 
studies and reflect the evidence on balance.

What are the desiderata of multi-document \emph{synthesis}?
First,  
summaries produced by models should be \textit{consistent} with the input data, with respect to the latent property of interest. 
In the case of Rotten Tomatoes, the sentiment of the summary should be in line with the aggregate sentiment expressed in the individual critic reviews.
A corollary to this is that models should be \textit{sensitive} to changes in the composition of inputs, e.g., removing most of the negative reviews from a set of inputs should yield a summary with a corresponding increase in the expressed sentiment. 

In this work we evaluate neural MDS models with respect to these criteria. 
To this end we use a meta-reviews dataset from Rotten Tomatoes \citep{rt-kaggle} 
and a dataset of systematic reviews (meta-analyses) summarizing the evidence about medical interventions \citep{Wallace2020GeneratingN}. 
For the former we probe the degree to which 
generated meta-review sentiment agrees with the expected aggregate sentiment score; for the latter we evaluate whether the generated summary indicates that the input evidence suggests, on balance, that the intervention under consideration was effective.

Our {\bf main contributions} are:
\begin{enumerate}
    \item To the best of our knowledge, this is the first work to investigate implicit \emph{synthesis} in summarization, and the degree to which modern models are capable of this.\footnote{\citet{shah-etal-2021-nutri} studies a low-resource health and nutrition setting, in which they extract relational tuples, apply a manual rule set for aggregation, and then generate a surface form following this result. See Section \ref{sec:related_work} for a discussion of Opinion Summarization work which considers synthesis as a \textit{target} but not measure of summarization performance.}

\item We show that ``off-the-shelf'' neural MDS models are somewhat inconsistent and insensitive with respect to performing synthesis in summarization.
\item We propose and evaluate a simple, general method of generating a diverse set of output candidates \citep{vijayakumar2016diverse} and then selecting from these based on agreement with an expected aggregate measure (based on inputs), with promising results. 
\end{enumerate}

\section{Synthesis and Summarization}

\begin{table*}
\small
    \centering
    \begin{tabular}{l lll||lll}
    & \multicolumn{3}{c}{Movie Reviews} & \multicolumn{3}{c}{Systematic Reviews} \\
    \hline
                                &    Train   &  Dev  & Test &    Train   &  Dev$^\dagger$  & Test\\\hline
    Number of metareviews                   &  7251      & 932   & 912 &  1675     & 360           & 397 \\
    Avg metareview length          &  32.0      & 32.6  & 32.4 &  101      & 107           & 111 \\
    Total number of inputs            &  195033    & 24336 & 24474 &  11054    & 1238          & 2669  \\
    Avg number of inputs          &  26.9      & 26.1  & 26.8 &  6.6      & 3.4           & 6.7\\
    Avg length of individual input      & 30.6       & 30.8  & 30.6& 475       & 379           & 449   \\
    Avg length of concatenated inputs   &  822       & 804   & 822  &  2641    & 1336          & 2544  \\
    Target Percent Positive             & 59.5  & 62.1 & 61.2  & 31.9      & 31.4          & 35.0 \\
    \hline
    \end{tabular}
    \caption{Dataset statistics for movie reviews (left) and systematic reviews (right). Number of meta-reviews, average meta-review length (tokens), input reviews per split, average number of inputs per instance, average total length of instance-inputs. For movie reviews, the target percent positive reports the fraction of metareviews with a positive sentiment; for systematic reviews this refers to the fraction of metareviews reporting a significant effect. $\dagger$ We subset the original dev set to instances of $\leq4k$ tokens (accommodating T5; other models can consume up to 16k).}
    \vspace{-1em}
    \label{tab:sentiment_stats}
    \label{tab:cochrane_dataset_stats}
\end{table*}

In standard multi-document summarization, we assume inputs $(X_i, y_i)$; $X_i=\{x_{i1}, ... , x_{i{|X_i|}}\}$. 
We then typically train a summarization model with parameters $\theta$, to consume $X_i$ and yield summaries $\hat{y}_i$ as similar as possible to targets $y_i$.
In a supervised setting, the standard objective estimates a $\theta$ to maximize target token log-probabilities. 
Assuming the input documents $x_{ij}$ in $X_i$ have been linearized (i.e., concatenated, with special tokens demarcating individual inputs) into an input string $x^{\oplus}_i$, this objective takes the form: $\sum_{t=1}^{|y_i|} \text{log } p_{\theta}(y_{it} | y_{i1}, ..., y_{i(t-1)},  x_i^{\oplus})$,
where $p_{\theta}$ is a probability assigned to the token at position $t$ in the target $y_{i}$ by a summarization model with parameters $\theta$. 
By myopically focusing on encouraging the model to produce tokens mimicking the targets, this objective aligns with standard (but flawed) measures of automated summary quality like ROUGE \citep{lin2004rouge}, which quantify $n$-gram overlap between targets $y_i$ and outputs $\hat{y}_i$.

We are interested in settings in which there is an additional, latent property $z_{ij}$ implicit in the constituent input texts $x_{ij}$. 
For example, $z_{ij}$ might reflect the sentiment in critique $j$ of the film indexed by $i$. 
Summaries should \emph{synthesize} this aspect, i.e., the generated summary $\hat{y}_i$ should implicitly convey an aggregated $z_{i}$ which reflects a synthesis or aggregation $G$ over $Z_i = \{z_{i1}, ... z_{i|X_i|}\}$. That is, we assume $z_{i} = G(Z_i)$ .
In both cases considered here---summaries of film critiques and synopses of clinical trials evidence---$G$ can reasonably be assumed to be a (weighted) mean, $G(Z_i) = \frac{1}{|X_i|} \sum_{j=1}^{|X_i|} \alpha_{ij} z_{ij}$. 
That is, summaries should roughly reflect the average sentiment and reported treatment effect in the cases of movie reviews and clinical trial reports, respectively.

We investigate the following questions. (1) Do model summaries $\hat{y}_i$ reflect the anticipated aggregate aspect of interest? That is, how well calibrated is the aspect communicated in the generated summary ($z_{i\hat{y}}$) compared to the expected $z_{i}$? 
(2) Do these same results apply to other (not solely transformer) MDS architectures?
(3) Can we \emph{improve} the ability of summarization models to synthesize by explicitly incorporating synthesis targets $z_{i}$ into the decoding process?

We propose a simple inference-time procedure to explicitly preference output candidates that align with the expected aggregate property of interest (e.g., average sentiment), and report promising results under both automatic and manual evaluation. %
This strategy naturally lends itself to \emph{cautious} summarization, i.e., approaches where the model can \emph{abstain} from generating an output if it does not produce any candidates that reflect the anticipated aggregate measure.

\begin{figure*}
    \centering
    \includegraphics[scale=0.40]{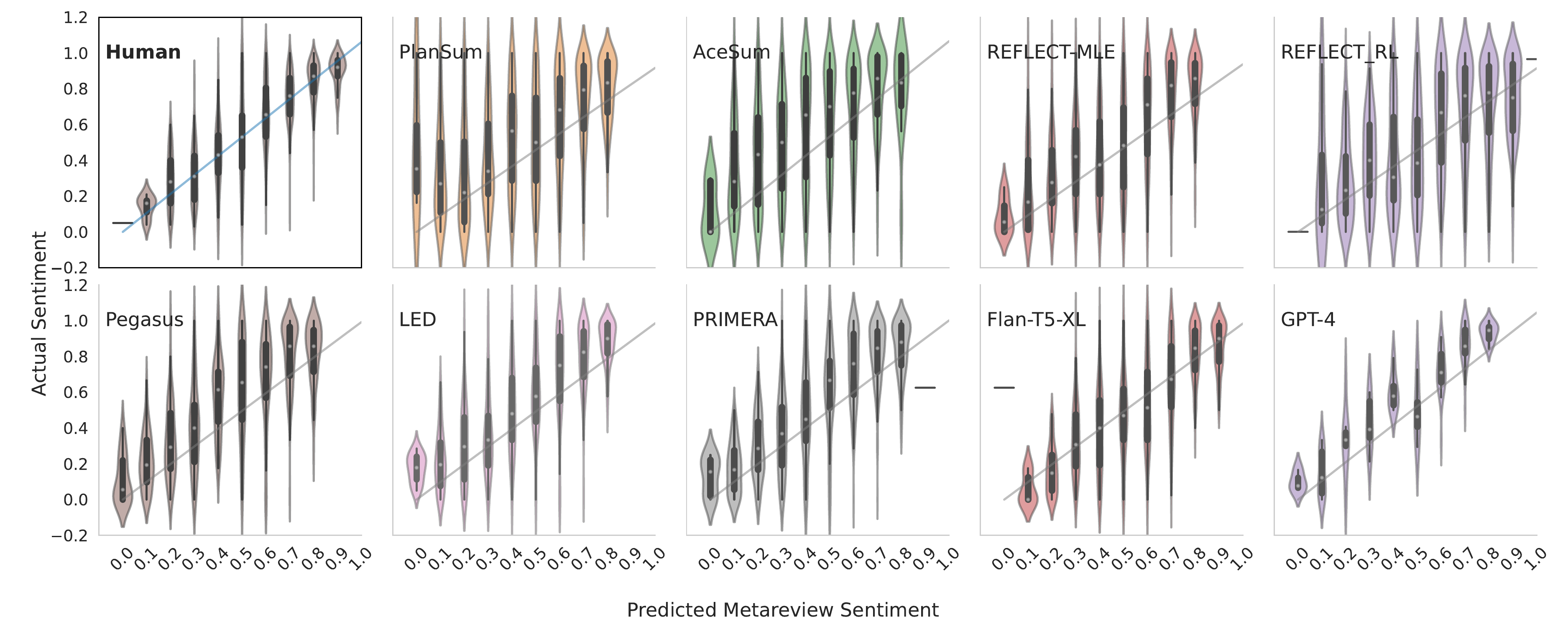}
    \caption{Movie Reviews: Actual vs. Predicted Sentiments on generated summaries. Human outputs replace LED (upper left) for comparison.
    }
    \label{fig:sentiment_measure_validation}
    \vspace{-1em}
\end{figure*}

\subsection{Movie Reviews} \label{sec:movies}

We first consider a dataset comprising movie reviews and associated meta-reviews summarizing these from Rotten Tomatoes. 
An in-house staffer (at Rotten Tomatoes) summarizes movie critic reviews\footnote{Written by designated ``top-critics'', critics recognized for quality and quantity of reviews in recognized publications} into meta-reviews \citep{NYT-RT}. 
These meta-reviews synthesize the input reviews, reflecting the aggregate critic reception of a film. 
Each meta-review is associated with a numerical ``Tomatometer'' score, which is an overall measure of the fraction of reviews that were positive (according to Rotten Tomatoes staffers) for the corresponding film (so here the target aggregation function $G$ would be this fraction).
The Rotten Tomatoes dataset we use comprises 9,095 movies with meta-reviews constructed from 244,000 individual reviews (Table \ref{tab:sentiment_stats}). 

\vspace{0.5em}
\noindent \textbf{Measuring sentiment in movie reviews.}
We need to measure the property of interest in texts; for this we use a \emph{measurement model} $g$---here we fine-tune a BERT model \citep{devlin-etal-2019-bert} using the continuous (fine-grained) sentiment targets provided in the SST dataset \citep{socher-etal-2013-recursive}.\footnote{We use the \textit{continuous} measurements from the original SST dataset, not the two or five class projections of those underlying measurements.}
We fine-tuned this model on the SST dataset for 3 epochs with a learning rate of 5e-5 using the {\tt Huggingface} library \citep{wolf-etal-2020-transformers} %
with no hyperparameter tuning.
While the raw text of the SST dataset is in-domain (i.e.,  movie reviews), the targets themselves are not.\footnote{SST is itself based on a collection of Rotten Tomatoes critic reviews \citep{pang-lee-2005-seeing}. 
We verified that the SST text fragments do not overlap with our target reviews by manually checking any (fragment, review) pair with substantial ($\geq75\%$) overlap for one quarter of all reviews.}
When applying this fine-tuned $g$ to the movie meta-reviews, we find a reasonably strong correlation between our sentiment estimates and the ``true'' meta-review sentiment (``Tomatometer'' score): The R$^2$ (centered) is 0.696, mean squared error (MSE) is 0.022, and Pearson's $r$ is 0.836 (Figure \ref{fig:sentiment_measure_validation}, upper left).\footnote{\label{measure_separation} In creating both synthesis measures $g$, we have \textit{isolated} them from our original datasets to not artificially favor human references as in-domain over machine generations.}

\subsection{Biomedical Systematic Reviews}
\label{sec:reviews}

Our second dataset is a collection of systematic reviews from the Cochrane Collaboration.\footnote{An international non-profit dedicated to helping healthcare providers make evidence-based decisions.} 
This dataset comprises roughly 2,600 systematic reviews summarizing a total of 16,500 clinical trials evaluating interventions in healthcare 
 (Tables \ref{table:systematic_review_example}, \ref{tab:cochrane_dataset_stats}). 
Each review includes a natural language summary and accompanying statistical meta-analysis results. 
The latter provides an aggregate statistical summary of the individual (study-level) data extracted from the trials included in each review. 
The natural language summary should accurately convey and contextualize the findings of the meta-analysis. 
Therefore, the (lack of) treatment efficacy communicated in a given summary should generally agree with the direction of the corresponding meta-analytic point estimate.

\vspace{0.5em}
\noindent \textbf{Measuring effects in evidence syntheses.} 
For systematic reviews of clinical trials, we resort to a less granular \emph{classification} model $g(x_{ij}), g(y_{i})$ which attempts to infer whether a text reports a significant result.
Specifically, we use {\tt RobotReviewer} \citep{marshall-etal-2017-automating,deyoung-etal-2020-evidence}. 
Given a narrative describing a clinical trial result (or a summary of trials), {\tt RobotReviewer} predicts whether the reported result indicates a significant effect of the treatment being investigated, or not.
We can compare this prediction to the ``truth'', which here is derived from the meta-analytic result (specifically by checking whether $p<0.05$). 
Applying this off-the-shelf model to the manually composed summaries accompanying the meta-analyses in our Cochrane set, we observe a macro-average F1 score of 0.577 and 68.6\% accuracy, %
providing a reasonable (if weak) measure for this task.\textsuperscript{\ref{measure_separation}}%

\section{Models}
\label{sec:models}

We %
evaluate a suite of transformer \citep{vaswani2017attention} summarization models: Pegasus \citep{Zhang2020PEGASUSPW}, Longformer \citep{Beltagy2020LongformerTL}, PRIMERA \citep{xiao-etal-2022-primera}, T5 \citep{Raffel2020ExploringTL} and Flan-T5 \citep{Chung2022ScalingIL}, and GPT-4 \citep{openai2023gpt4}.
For each trainable transformer model and dataset we performed a hyperparameter search over learning rates and training steps (retaining most parameter defaults). We train with an effective batch size of 16 and floating point 16\footnote{Flan-T5-Large and -XL used BF16 for speed} precision on an NVIDIA RTX-8000 GPU (due to data size we can fit only a single instance in memory at a time for some models, and must use gradient accumulation).

Models were fine-tuned using the Adam optimizer \cite{Kingma2014AdamAM}, except Pegasus which was fine-tuned with Adafactor \cite{pmlr-v80-shazeer18a},\footnote{In larger Flan-T5 models we experimented with both optimizers; differences in ROUGE1 performance were small.} across several learning rates (1e-4, 1e-5, 1e-6), for up to 20k training steps. The best model was selected based on ROUGE-1 performance on the validation set.\footnote{\href{https://github.com/jayded/MDSSynthesis}{https://github.com/jayded/MDSSynthesis}}
PRIMERA was designed and pre-trained specifically for multi-document summarization. %
Though not explicitly designed as multi-document summarization models,  both Pegasus \cite{Zhang2020PEGASUSPW} and %
T5 \citep{amplayo-etal-2021-aspect}
have been used on multi-document tasks, while 
Longformer has been used for a related multi-document summarization task \citep{deyoung-etal-2021-ms}.

For GPT-4 (-0613) we use system prompt \textit{You are a professional movie critic. Your job is to provide an opinionated summary of a movie, in your own words. You will have access other critics'   opinions of the movie.} and assistant prompt \textit{For movie \{movie\}, other critics have written: \{reviews\}. In your own words, please produce an opinionated summary of \{movie\}.}, providing a one-shot example. %
For systematic reviews, we used the system prompt \textit{You are a systematic reviewing expert. Your job is to read randomized control trial reports and assist a medical researcher. You will aid in drafting systematic reviews.} with assistant prompt:\  \textit{Please provide a draft systematic review for the studies below: \{studies\}. Start with the conclusions of the review only, a more detailed analysis will happen later},
 again providing a single shot example.
 
 As %
 it is not the focus of our work here, we did not extensively tune these  prompts.
 We inspected outputs over five training instances when developing prompts for both movies and systematic reviews datasets. When designing movie review prompts, we iterated through first asking the model to \textit{summarize} the reviews (yielding a summary of each review instead of an aggregate), %
 then telling the model to use the same language as the reviews (with effectively the same result), then providing a single example (yielding some improvement), then demanding an \textit{opinionated summary} (again with some improvement), and finally telling the model to use its own words (yielding the prompt above and experiments below). 
 For the systematic review prompt, we first we asked for a draft review (the model provided an entire draft), then we specified conclusions only (we received an abbreviated abstract), then we specified a conclusions \textit{section} (we received a less abbreviated abstract), and, finally,  adding an in-context example. 
 We also explored asking for a high level \textit{summary} (rather than systematic review) of the input studies; %
 and with prompts providing intervention and outcome information to the model and asking for a draft of the review.  %

Beyond transformers, we consider models from the opinion summarization and content aggregation literature: PlanSum \cite{Amplayo2020UnsupervisedOS}, QT \cite{angelidis-etal-2021-extractive}, AceSum \cite{amplayo-etal-2021-aspect}, and REFLECT \cite{song-etal-2022-improving}.\footnote{We considered HierSumm \cite{liu-lapata-2019-hierarchical}, but excluded it for extreme degeneration while decoding. We excluded Hercules \cite{hosking-etal-2023-attributable} as the software was not adaptable to our tasks.} %
PlanSum \cite{Amplayo2020UnsupervisedOS} learns a (disentangled) sentiment and aspect model, and augments an LSTM equipped with an attention-copy mechanism \cite{Bahdanau2014NeuralMT,NIPS2015_29921001} with this information as a decoder.

QT \cite{angelidis-etal-2021-extractive} learns a quantized embedding for each model input via an auto-encoder, then finds representative input sentences (via clustering and assignment) to use as summaries. We include QT\footnote{For movie reviews, where targets can appear similar to inputs in length and content, as opposed to systematic reviews (for which we do not evaluate QT), where the target prose differs substantially from its inputs.} as an extractive model. 
AceSum \cite{amplayo-etal-2021-aspect} adopts a hierarchical approach, representing each input document as sentences pooled over individual inputs, and passing this representation to a transformer (T5; \citealp{Raffel2020ExploringTL}), along with specific aspect or general codeword tokens and vocabulary embeddings, \textit{controlling} what type of summary to produce (we focus on the \textit{general} case).
REFLECT \cite{song-etal-2022-improving} takes the hierarchical approach one step further, with a sentence level extraction phase (using aggregated token representations) followed by an abstraction phase (BART; \citealp{lewis-etal-2020-bart}), trained via standard MLE and via a reinforcement learning credit aware self-critic method \cite{8099614}.
For all models we largely retained the original hyperparameters, with modifications to increase sequence lengths and decrease aspects (these models were developed around \textit{aspect} summarization).

\section{Experiments}

\begin{table}
    \centering
    \small
    \begin{tabular}{llll}\hline
        {} & R$^2$ & PCC & R1 \\ \hline
        QT & 0.592 & 0.788 & 0.122 \\ 
        PlanSum & 0.245 & 0.510 & 0.160 \\ 
        AceSum & 0.158 & 0.439 & 0.176 \\ 
        REFLECT$^{\text{MLE}}$ & 0.430 & 0.657 & 0.241 \\ 
        REFLECT$^{\text{RL}}$ & 0.225 & 0.507 & 0.218 \\ 
        Pegasus & 0.530 & 0.730 & 0.245 \\ 
        LED & 0.551 & 0.742 & 0.242 \\ 
        PRIMERA & 0.608 & 0.780 & 0.254 \\ 
        T5-Small & 0.441 & 0.669 & 0.234 \\ 
        T5-Base & 0.516 & 0.720 & 0.253 \\ 
        Flan-T5-S & 0.412 & 0.647 & 0.237 \\ 
        Flan-T5-B & 0.597 & 0.774 & 0.247 \\ 
        Flan-T5-L & 0.484 & 0.696 & 0.248 \\ 
        Flan-T5-XL & 0.611 & 0.783 &\textbf{0.262} \\ 
        GPT-4 & \textbf{0.808} & \textbf{0.900} & 0.166 \\ 
        Reference & 0.697 & 0.836 & ~ \\ \hline
    \end{tabular}      
    \caption{Synthesis results for \textbf{Movie reviews}: correlations (R$^2$, Pearson's $r$) between sentiment measured in model outputs %
      and Tomatometer Ratings. R1 is ROUGE1.
    } 
    \label{tab:sentiment_models_synthesis_score}
    \vspace{-1em}
\end{table}

\subsection{%
Do Summarization Models Synthesize?}

We report sentiment performance for all models in Table \ref{tab:sentiment_models_synthesis_score}.
These metrics quantify the strength of the relationship between (a) the continuous sentiment inferred (via our text regression 
measurement model $g$) over model generated or reference summaries and (b) the reference sentiment (Tomatometer) score.

Save for GPT-4, correlations between the sentiment measured in generated outputs and Tomatometer scores are considerably lower than that between the same measurement over human-composed summaries and said score. 
This implies that human authors tend to do a better job of synthesis than models when composing summaries. 
GPT-4 seems %
performs especially well here; we are not entirely sure why, but it may owe to the differences in lengths of outputs (133 tokens on average vs. 31 for reference summaries).

For systematic reviews (Section \ref{sec:reviews}), the measurement model $g$ attempts to infer whether a text reports a significant treatment effect; we compare this against the $p$-value from the corresponding statistical meta-analysis.
This permits a coarse assessment of synthesis, as we are unable to measure correlations.
Instead we report classification metrics describing how often the effect significance inferred from a summary (generated or manually written) matches the ground truth derived from the meta-analysis (Table \ref{tab:cochrane_review_significance_f1s_dev_set_sample}). 
The results are qualitatively similar to the sentiment case, in that the humans appear to do a better job of synthesis---as best we can measure, the  significance reported in their summaries better aligns with the statistical results than in model generated summaries. GPT-4 is again an exception,  slightly \emph{outperforming} human results on this metric, which may owe to its formulaic generation featuring strong, direct, clear initial statements of treatment effectiveness.

\begin{figure*}[t]
    \centering
    \begin{subfigure}{.48\textwidth}
        \centering
        \includegraphics[width=\textwidth]{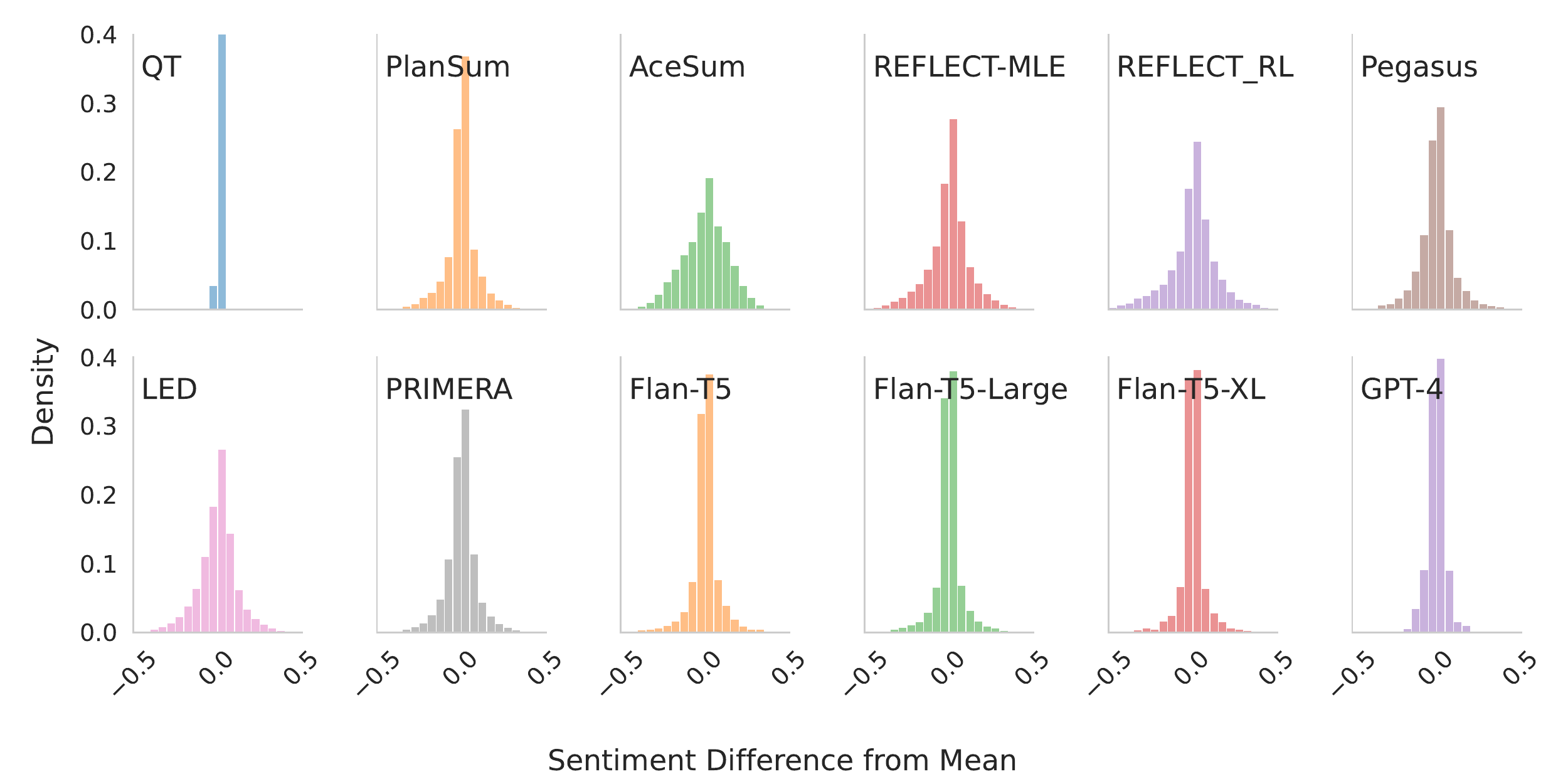}    
    \end{subfigure}
    \begin{subfigure}{.48\textwidth}
        \centering
        \includegraphics[width=\textwidth]{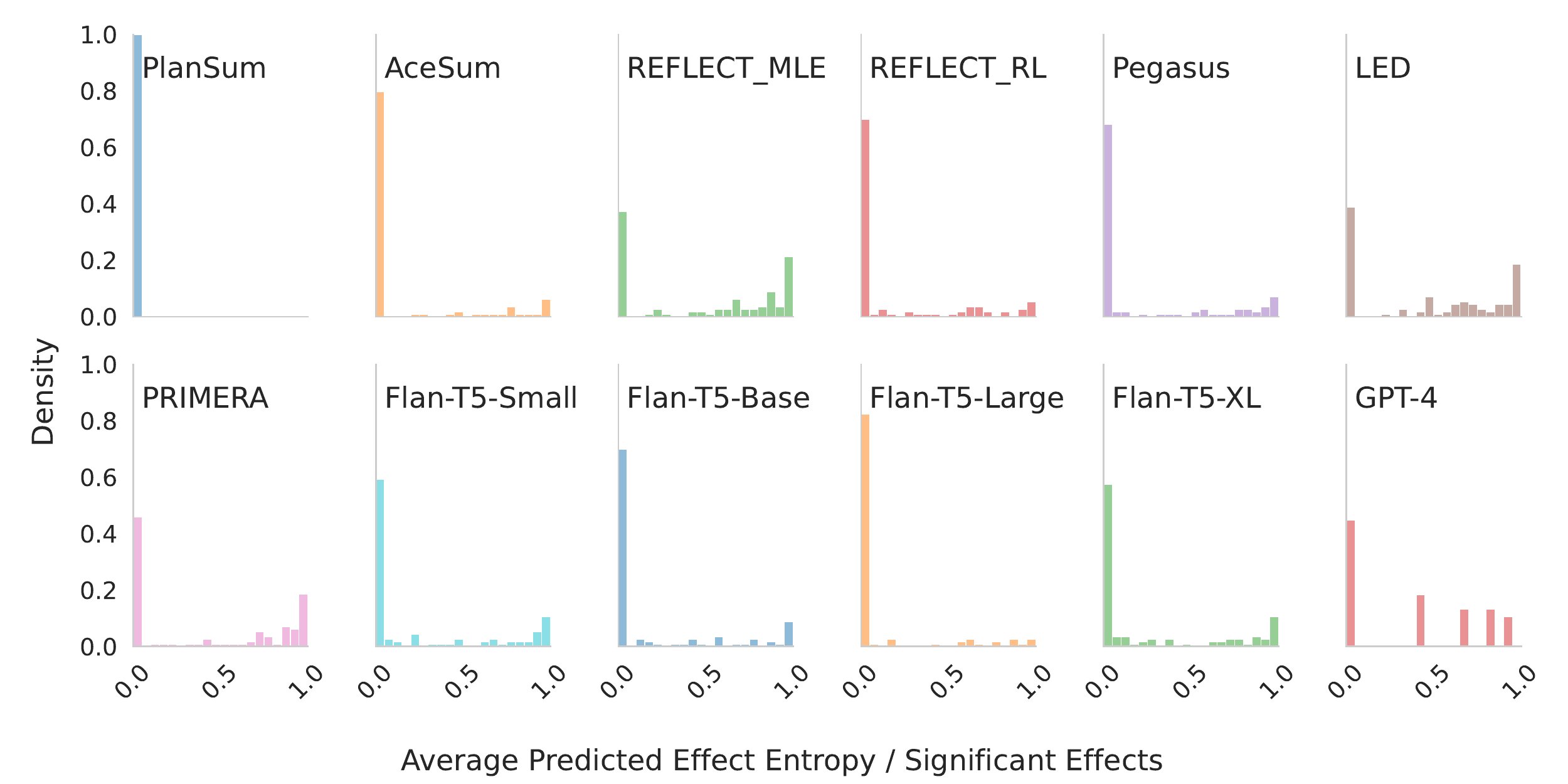}
    \end{subfigure}
    \caption{The spread of sentiment/treatment effect measured in outputs produced from  permuted input orderings. Left: Movie review sentiment. Right: Systematic review significance prediction entropy (0 indicates order insensitivity) %
    on the subset of reviews that report \emph{significant} effects.}%
    \label{fig:shuffling_differences_hist}
    \label{fig:cochrane_predicted_effect_results}
\end{figure*}

\begin{table}
\centering
\small
                \begin{tabular}{lrrrr}\hline
        {}              &  F1               & Acc   &  R1 \\\hline
        PlanSum         &  0.414            & 0.683 & 0.177 \\
        AceSum          &  0.532            & 0.550 & 0.151 \\
        REFLECT$^{\text{MLE}}$     &  0.532 & 0.639 & 0.271 \\
        REFLECT$^{\text{RL}}$      &  0.505 & 0.683 & 0.199 \\
        Pegasus         &  0.568            & 0.714 & 0.212 \\
        LED             &  0.490            & 0.631 & 0.259 \\
        PRIMERA         &  0.526            & 0.644 & 0.253 \\
        T5-Small        &  0.540            & 0.600 & 0.205 \\
        T5-Base         &  0.521            & 0.628 & 0.206 \\
        Flan-T5-Small   &  0.548            & 0.583 & 0.081 \\
        Flan-T5-Base    &  0.538            & 0.683 & 0.194 \\
        Flan-T5-L       &  0.556            & \textbf{0.692} & 0.218 \\
        Flan-T5-XL      &  0.487            & 0.608 & 0.268 \\
        GPT-4           &  \textbf{0.628}   & 0.640 & \textbf{0.273} \\
        Reference       &  {0.577}          & 0.686 & \\     
        \hline
        \end{tabular}

    \caption{Synthesis results for \textbf{Systematic reviews}: Macro-averaged F1s and accuracies ({\tt RobotReviewer} predictions over model outputs vs. reference meta-analysis results). 
    } 

    \label{tab:cochrane_review_significance_f1s_dev_set_sample}
    \vspace{-1em}
\end{table}

\begin{figure*}[t]
\centering
\begin{subfigure}{.48\textwidth}
    \centering
    \includegraphics[width=\textwidth]{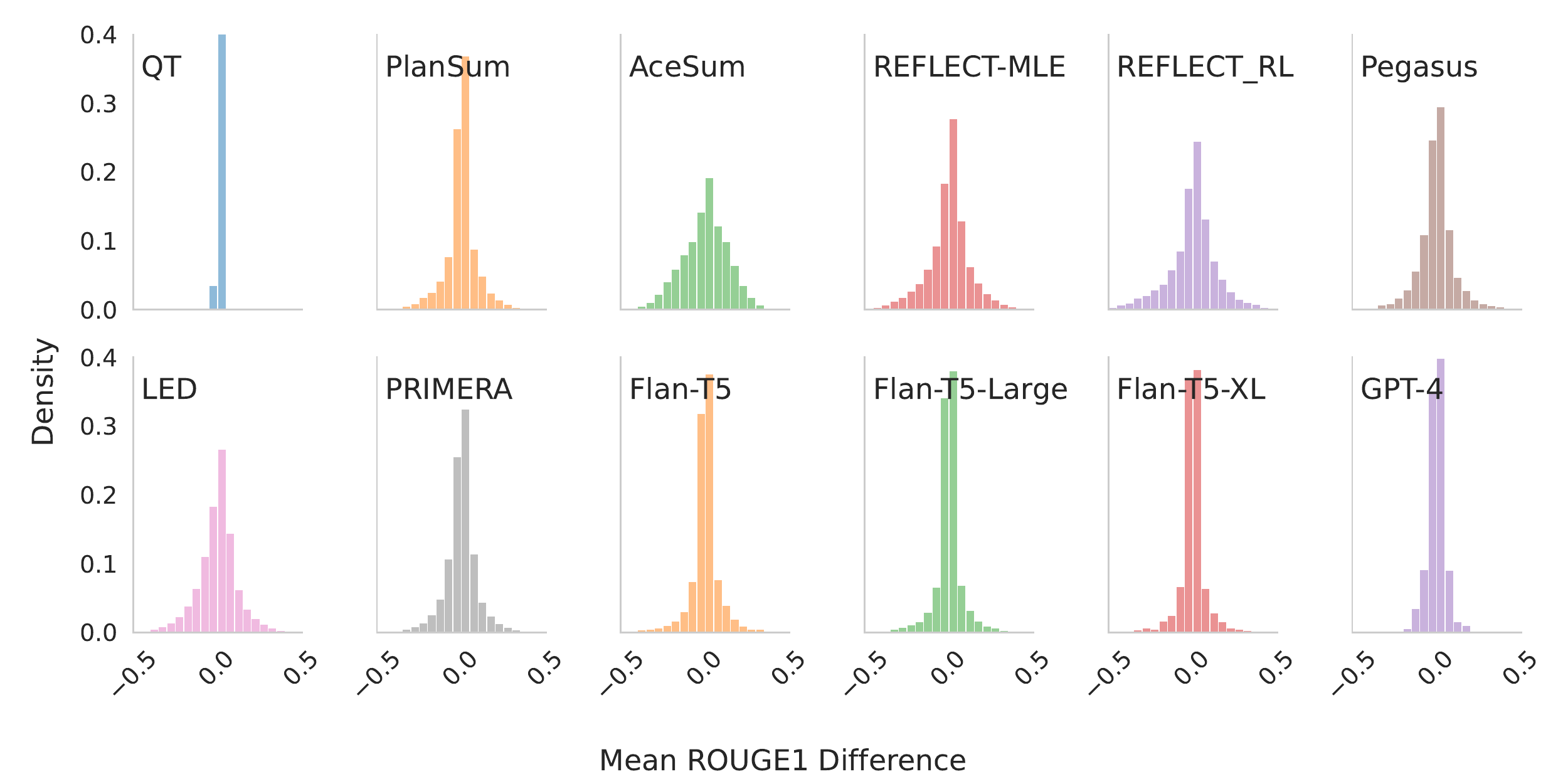}
\end{subfigure}
\begin{subfigure}{.48\textwidth}
        \centering
        \includegraphics[width=\textwidth]{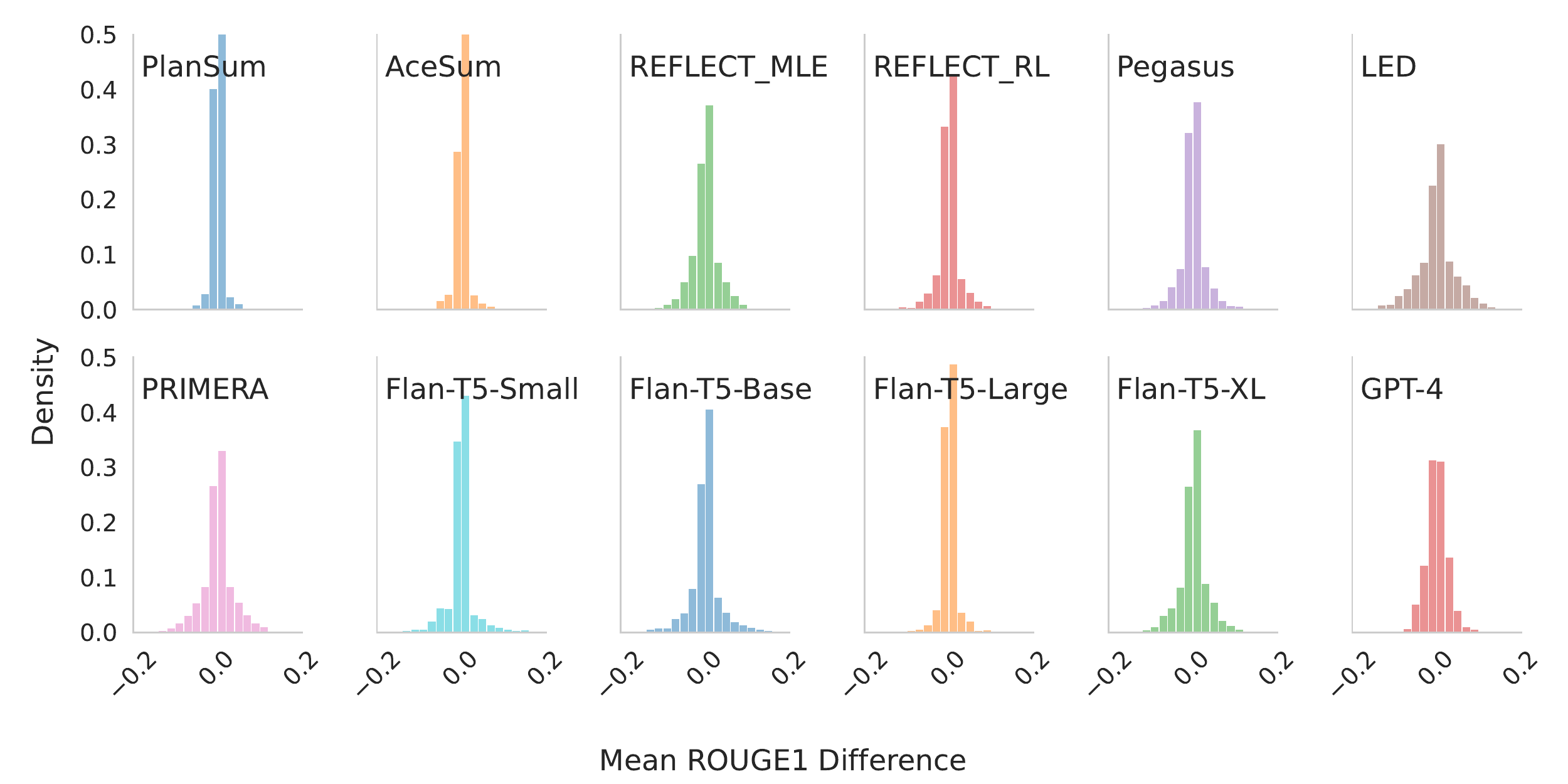}
\end{subfigure}
\caption{ROUGE1 deltas from instance means for movie reviews (left) and systematic reviews (right).}
\label{fig:movies_sentiment_rouge_stddev_histogram}
\vspace{-0.5em}
\end{figure*}

\subsection{Sensitivity to Input Ordering} \label{sec:shuffling}

Synthesis of inputs should be invariant to ordering (e.g., critic consensus on a film does not depend on the order in which one reads the reviews). 
Here we evaluate if models are sensitive to input ordering with respect to the synthesized aspect of interest ($z_{i\hat{y}}$). 
Specifically, let $X_i = \{x_{i1}, ... , x_{i{|X_i|}}\}$ denote an arbitrary ordering of inputs in the linearized version $x^{\oplus}_i$. 
This ordering should not affect the aggregate aspect $z_{i\hat{y}}$ in the summary. 

To evaluate if models realize this invariance, we permute the instance $i$ inputs $X_i$ (and, consequently, the linearized $x^{\oplus}_i$) one hundred times,\footnote{As a cost saving measure, we sample ten times for GPT, over one hunded different inputs instead of the full development set. Our experiments cost approximately \$500 to run.}
randomizing input orderings. 
For each such permutation $\tilde{X}_i$ (and associated $\tilde{x}^{\oplus}_i$), we generate a summary $\hat{y}_i$ and estimate of the resultant aspect $\tilde{z}_{i\hat{y}}$, using the corresponding measurement model.
By repeating this process for each instance $i$, we can construct an empirical distribution over $\tilde{z}_{i\hat{y}}$'s under different random orderings.

\textbf{Movie reviews.} We zero-mean the $\tilde{z}_{i\hat{y}}$'s inferred over each instance, and combine the distributions from all instances into a histogram (Figure \ref{fig:shuffling_differences_hist}). 
This shows the spread of sentiments inferred over outputs under random input orderings minus the corresponding instance mean sentiment.  
Were a model completely invariant to ordering, the empirical distribution over these differences would collapse to 0.
Instead, we observe a relatively wide spread in sentiment measured over outputs generated from different permutations, indicating a counter-intuitive sensitivity to orderings. (Interestingly, Figure \ref{fig:movies_sentiment_rouge_stddev_histogram}---provided for comparison---suggests such permutations also affect ROUGE; we do not explore this aspect further here.) %

\textbf{Systematic reviews}. For each $X_i$ we have 100 order permutations and associated summaries; we infer whether these report \emph{significant results} or not, and record the fraction that do ($p_i$).
If models were invariant to ordering, this fraction would always be 0 or 1. 
Values in-between suggest the model flips the report conclusion as a result of different input orderings. 
Figure \ref{fig:cochrane_predicted_effect_results} (right) shows a histogram of entropies over $p_i$, computed over the subset of examples where the associated meta-analysis indicates a significant effect.
Densities away from zero indicate sensitivity to ordering. QT, PlanSum, and GPT-4 all have a smaller spread than the other models --- QT because it is order insensisitive by construction, PlanSum similarly (but not entirely), and GPT-4 due to overall quality performance. 
We note that sensitivity is clearly an undesirable trait (\textit{any} spread is undesirable), but this may trade off against other metrics of interest.

\vspace{-0.5em}

\subsection{Sensitivity to Input Composition} \label{sec:sensitivity}

\begin{figure*}
    \centering
    \includegraphics[scale=0.53]{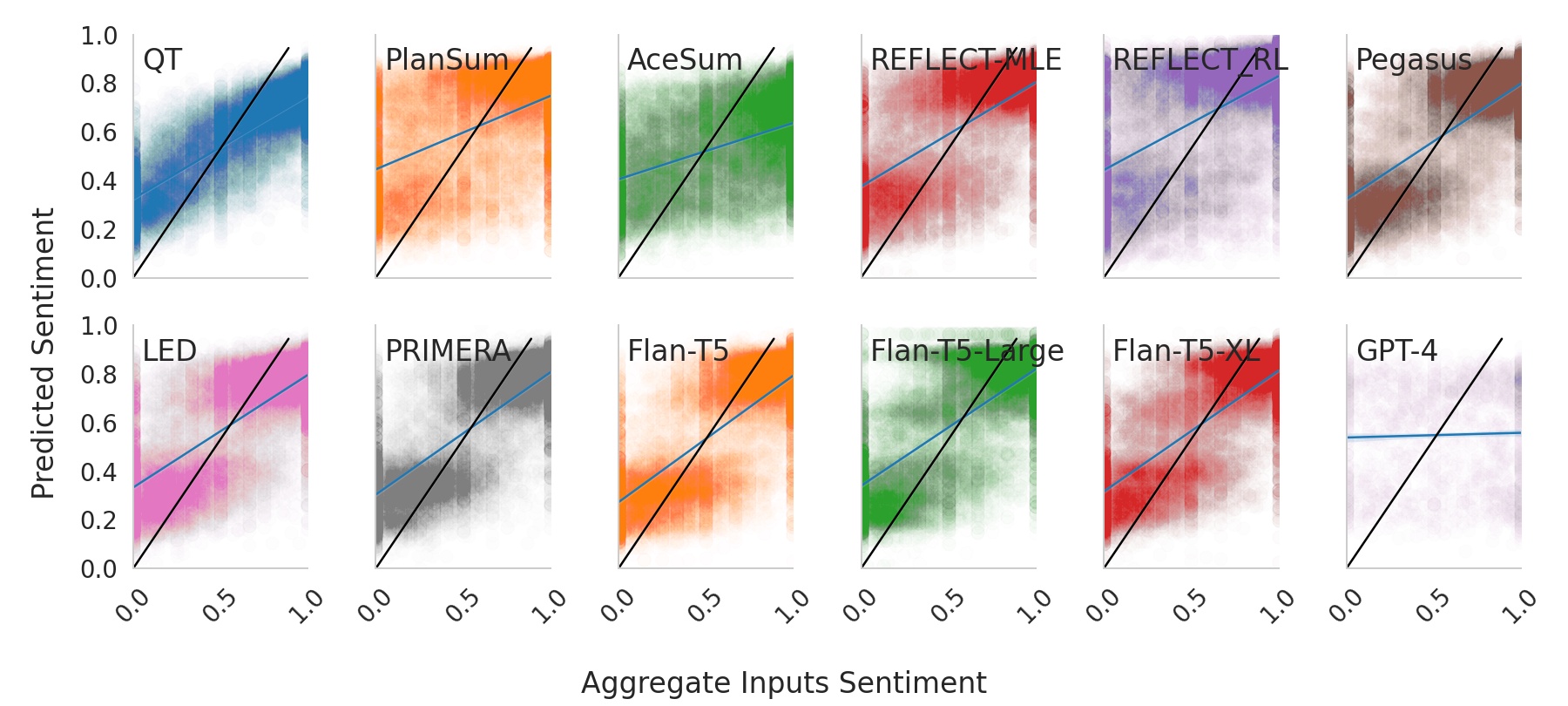}
    \caption{Model sensitivity to manipulated input sentiment composition. Intensity patterns indicate that models oscillate between low and high sentiments in outputs, and are not responsive to subtler shifts in input sentiment. We show a model regression (blue) and the reference sensitivity regression (black).}
    \label{fig:movies_sensitivity_derivates}
    \vspace{-0.5em}
\end{figure*}

Synthesis models should be responsive to changes in the distribution of the attribute to be synthesized in the input composition: If we increase the ratio of positive to negative reviews in an input set, we would anticipate a concomitant change in the sentiment communicated in the meta-review $z_{i\hat{y}}$.
To assess if models meet this synthesis desiderata, we manipulate model inputs $X_i$ in such a way to induce an expected change in the target measure $z_{i\hat{y}}$; we then measure if the output yields a summary that aligns with this expected change.

\textbf{Movie reviews}. We manipulate the ratio of positive to negative reviews and observe the resultant change in the property of interest latent in the corresponding output. 
We take movies with mixed reviews, and delete 10\%, 20\%, 30\%, ..., 100\% of the positive inputs, retaining the negative inputs; we then repeat the process but instead remove negative inputs. 
For each of these permutations, we measure the input sentiment, the meta-review sentiment, and how well they correlate (Table \ref{tab:movies_correlation_sensitivity}).

\begin{table}
\small
\centering
    \begin{tabular}{lll}%
    \hline
  {}  & R$^2$ & PCC   \\\hline %
QT & 0.634 & 0.796 \\
PlanSum & 0.249 & 0.499 \\
AceSum & 0.177 & 0.420  \\
REFLECT$^{\text{MLE}}$ & 0.439 & 0.663 \\
REFLECT$^{\text{RL}}$ & 0.294 & 0.542 \\
Pegasus & 0.499 & 0.706  \\
LED & 0.524 & 0.724  \\
PRIMERA & 0.572 & 0.756  \\
T5-Small & 0.447 & 0.668 \\
T5-Base & 0.481 & 0.694  \\
Flan-T5-Small & 0.393 & 0.627  \\
Flan-T5-Base & 0.556 & 0.746  \\
Flan-T5-Large & 0.490 & 0.700  \\
Flan-T5-XL & 0.551 & 0.742  \\
GPT-4 & 0.457 & 0.677  \\\hline
    \end{tabular}
    \caption{{\bf Movie reviews} Correlations between subsampled inputs and generations.
    }
    \label{tab:movies_correlation_sensitivity}
    \vspace{-1em}
\end{table}

\begin{table}
\vspace{-1.7em}
\small
    \centering
    \begin{tabular}{lll}
    \hline
        {} & F1 & Acc  \\ \hline
        PlanSum & 0.442 & 0.741 \\
        AceSum & 0.454 & 0.504  \\
        REFLECT$^{\text{MLE}}$ & 0.471 & 0.583  \\
        REFLECT$^{\text{RL}}$ & 0.445 & 0.689  \\
        Pegasus & 0.452 & 0.680  \\
        LED & 0.510 & 0.684  \\
        PRIMERA & 0.533 & 0.675  \\
        T5-Small & 0.560 & 0.618 \\
        T5-Base & 0.469 & 0.658  \\
        Flan-T5-Small & 0.430 & 0.500  \\
        Flan-T5-Base & 0.482 & 0.680  \\
        Flan-T5-Large &  0.435 & 0.693 \\
        Flan-T5-XL & 0.464  & 0.649 \\
        GPT-4 & 0.511 & 0.530\\ \hline
    \end{tabular}
    \caption{{\bf Systematic reviews}: Classification performance for subsampled inputs and generations. See Figure \ref{fig:cochrane_subsample_entropy} for a visualization of classification \textit{distribution}, analogous to Figure \ref{fig:movies_sensitivity_derivates} for movies. 
    }
    \label{tab:cochrane_reviews_sensitivity}
    
\end{table}

\begin{figure}
    \centering
    \includegraphics[width=.45\textwidth]{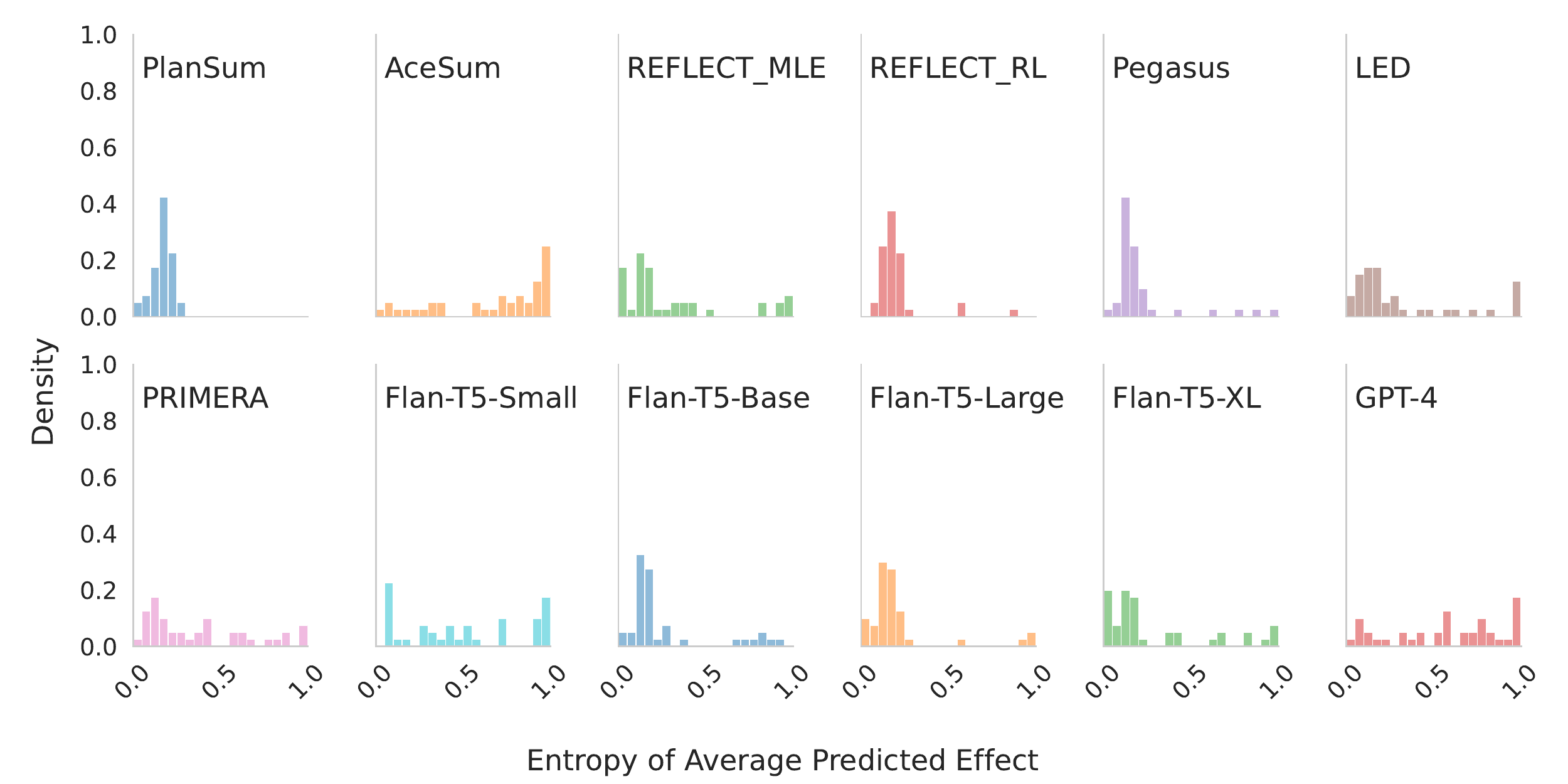}
    \caption{\textbf{Systematic Reviews.} A histogram of entropies for the \textit{subsampled} review classifications (where the ground truth is positive).}
    \label{fig:cochrane_subsample_entropy}
\end{figure}

Figure \ref{fig:movies_sensitivity_derivates} plots the relationship between the fraction of positive reviews in the (manipulated) input sets and the granular sentiment score inferred over the resultant outputs.  
The models are generally undersensitive to changes in their input: rather than having a change in meta-review sentiment equivalent in size to changes in input sentiment (a slope of 1, as we observe when we fit a model to the human written summaries).
Models tend to have trouble changing their sentiment, and require a large change in input distribution to substantially change the sentiment communicated in the output.

\textbf{Systematic Reviews}. To measure sensitivity to changes in input composition, we manipulate inputs $X_i$ such that the meta-analysis result (target $z_{i\hat{y}}$) flips from a significant effect to no effect, or from no effect to an effect (Table \ref{tab:cochrane_reviews_sensitivity}, Fig. \ref{fig:cochrane_subsample_entropy}). 
We first take a subset of the reviews that have conflicting evidence (139 unique reviews).
We then order inputs in these by (weighted) effect sizes,\footnote{In fixed effects meta-analysis the weights are inverse variances associated with study-level effect estimates.} and remove subsets which ought to flip the significance result of a subsequent meta-analysis. The surface level results (Table \ref{tab:cochrane_reviews_sensitivity}) show little difference from earlier results (i.e. the $\Delta$ values are approximately comparable to Table \ref{tab:cochrane_review_significance_f1s_dev_set_sample}), but our classification results become substantially noisier (Figure \ref{fig:cochrane_subsample_entropy}). We speculate that models are picking up on some uncertainty from the change in overall meta-analysis but overall fail to capture that detail in their outputs. Even if the models reflect uncertainty due to the \textit{strength} of the change (desirable!) this is still \textit{incorrect} as the finding has changed.

{\textbf{Result.} In both the case of the Movie Reviews and the Systematic Reviews, we see a substantial drop in performance from the base review results (reported in Tables \ref{tab:sentiment_models_synthesis_score},\ref{tab:cochrane_review_significance_f1s_dev_set_sample}). We can only speculate as to the cause of this. 
Perhaps this indicates memorization of original targets in pre-training, or maybe removing strong (positive or negative) reviews hampers performance. 

\section{Improving Synthesis in Summarization}
\label{section:improving}

We propose a straightforward post-hoc approach to improving the synthesis performed by multi-document summarization models: (1) Generate an explicitly \emph{diverse} set of output candidates; (2) Select from these as the final output the candidate that best agrees with the expected synthesis result (as predicted by an external model).\footnote{Oved and Levy (\citeyear{oved-levy-2021-pass}) explore a related generate-then-select approach for creating \textit{plausible} product reviews. We experimented with an additional decoding method: constrain beam search by restricting candidate productions $p_{\theta}(y_{i,t} | y_{i,1..t-1},  x_i^{\oplus})$ such that the target attribute $z_i$ is less than some $\epsilon$: $\left|g(\hat{y}_{i,1,..,t}) - z_i\right| < \epsilon$. We elide these results here as they were often disfluent.} %

\begin{figure*}[ht]
    \centering
    \includegraphics[scale=0.375]{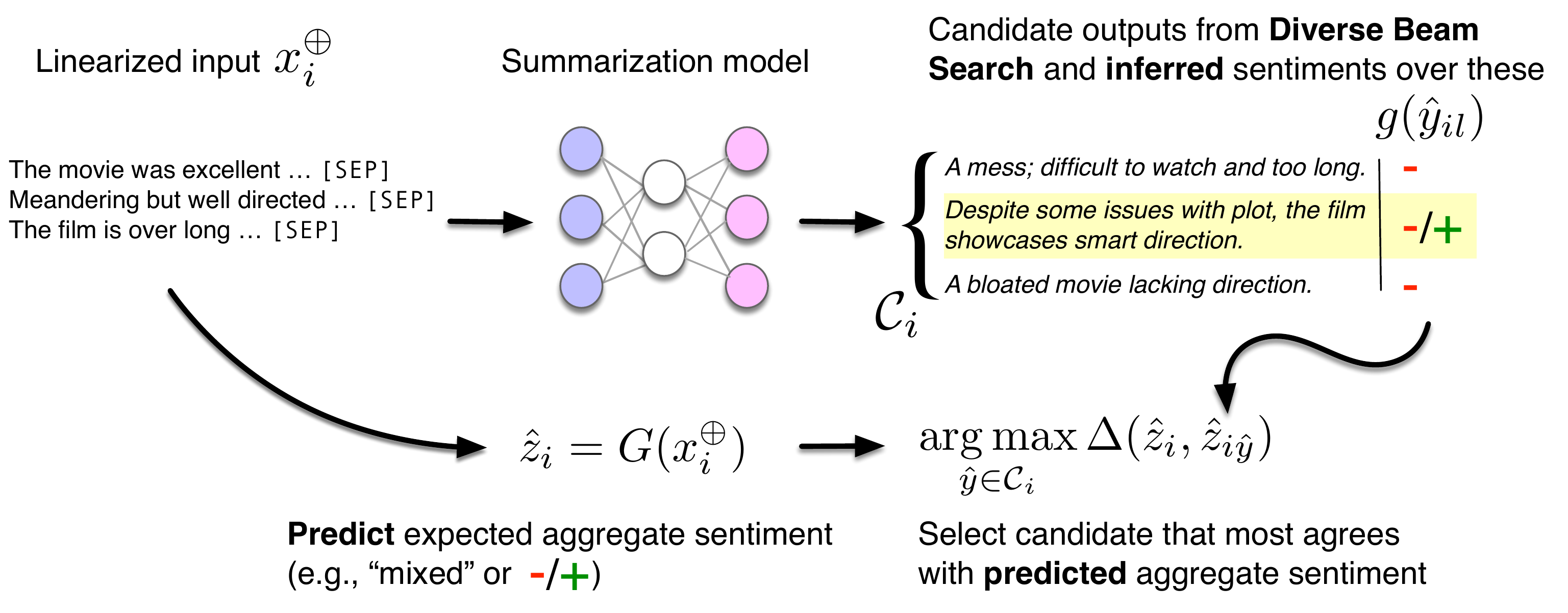} \vspace{-0.25em}
    \caption{Our proposed strategy to improve synthesis. We generate an diverse set of output candidates \citep{vijayakumar2016diverse} and then select the text that best agrees with the \emph{predicted} aggregate property of interest (here, sentiment). We can also \emph{abstain} when the model fails to yield an appropriate output.}
    \label{fig:decode-for-synthesis}
\end{figure*}

For (1), we rely on an existing technique for generating diverse outputs $\mathcal{C}_i$ from input $x^{\oplus}_i$: \emph{Diverse Beam Search} (DBS) \citep{vijayakumar2016diverse}.
This method modifies standard beam search to maintain multiple \emph{groups} of beams.
During decoding, a term is added to the next-token log probabilities, penalizing production of strings similar to candidates in \emph{other} groups.\footnote{This penalty requires a hyperparameter $\lambda$ that encodes the relative importance of diversity; we use $\lambda$=0.5.
To enable fair comparison with standard beam search (5 beams, in all experiments), we used 5 groups, 1 beam per group.
We exclude QT as it is an extractive model, and PlanSum as it does not readily support diverse beach search.
For AceSum and REFLECT we modify these codebases to use the diverse beam search implementation from HuggingFace. For GPT-4 we sample five responses with a temperature of 0.6.}

\begin{table*}
\footnotesize
\begin{tabular}{ p{5.25em} llllll||llllll}\hline
 {}& \multicolumn{6}{c||}{Approximate Selection} & \multicolumn{6}{c}{Oracle Selection} \\
   {} & R$^2$ & $\Delta$ & PCC & $\Delta$ & R1 & $\Delta$ & R$^2$ & $\Delta$ & PCC & $\Delta$ & R1 & $\Delta$ \\ \hline
        AceSum & 0.566 & \textbf{0.408} & 0.769 & \textbf{0.330} & 0.162 & -0.014 & 0.723 & \textbf{0.565} & 0.861 & 0.422 & 0.162 & -0.014 \\ 
        REFLECT$^{\text{MLE}}$ & 0.658 & 0.228 & 0.825 & 0.168 & 0.241 & \phantom{-}0.000 & 0.791 & 0.361 & 0.895 & 0.238 & 0.240 & -0.001 \\ 
        REFLECT$^{\text{RL}}$ & 0.491 & 0.266 & 0.702 & 0.195 & 0.220 & \phantom{-}0.002 & 0.576 & 0.351 & 0.759 & 0.252 & 0.219 & \phantom{-}0.001 \\ 
        Pegasus & 0.694 & 0.164 & 0.835 & 0.105 & 0.229 & -0.016 & 0.799 & 0.269 & 0.894 & 0.164 & 0.232 & -0.013 \\ 
        LED & 0.656 & 0.105 & 0.821 & 0.079 & 0.229 & -0.013 & 0.763 & 0.212 & 0.878 & 0.136 & 0.227 & -0.015 \\ 
        PRIMERA & 0.749 & 0.141 & 0.880 & 0.100 & 0.240 & -0.014 & 0.890 & 0.282 & 0.948 & 0.168 & 0.240 & -0.014 \\ 
        T5-Small & 0.692 & 0.251 & 0.846 & 0.177 & 0.225 & -0.009 & 0.827 & 0.386 & 0.913 & 0.244 & 0.226 & -0.008 \\ 
        T5-Base & 0.721 & 0.205 & 0.856 & 0.136 & 0.231 & -0.022 & 0.876 & 0.360 & 0.938 & 0.218 & 0.230 & -0.023 \\ 
        Flan-T5-S & 0.698 & 0.286 & 0.837 & 0.190 & 0.219 & -0.018 & 0.832 & 0.420 & 0.912 & 0.265 & 0.218 & -0.019 \\ 
        Flan-T5-B & 0.732 & 0.135 & 0.863 & 0.089 & 0.225 & -0.022 & 0.863 & 0.266 & 0.930 & 0.156 & 0.225 & -0.022 \\ 
        Flan-T5-L & 0.732 & 0.248 & 0.866 & 0.170 & 0.243 & -0.005 & 0.875 & 0.391 & 0.937 & 0.241 & 0.244 & -0.004 \\ 
        Flan-T5-XL & 0.769 & 0.158 & 0.888 & 0.105 & \textbf{0.250} & -0.012 & 0.900 & 0.289 & 0.950 & 0.167 & \textbf{0.248} & -0.014 \\ 
        GPT-4 & \textbf{0.814} & 0.006 & \textbf{0.924} & 0.024 & 0.159 & -0.007 & \textbf{0.914} & 0.106 & \textbf{0.963} & 0.063 & 0.164 & -0.002 \\ 
        Reference & 0.697 & {} & 0.836 & {} & {} & {} & 0.697 & {} & 0.836 & {} & {} & {} \\ 

        \hline
    \end{tabular}
     
    \caption{\normalsize\textbf{Movie Reviews}: Generate diverse meta-reviews and select from them using an approximate (left) or oracle (right) target sentiment. Performance improves on every measure except ROUGE-1. $\Delta$s compare the metric to their left with the results reported in Table \ref{tab:sentiment_models_synthesis_score}.
    } %
    \label{tab:sentiment_diversity_extracted_truth}
    \label{tab:sentiment_diversity_ground_truth}
\end{table*}

In (2) we would like to select the output that best synthesizes the property of interest; this requires an approach to specify what we \emph{expect} the synthesized property be, given the inputs.
For example, if we know the sentiment scores associated with input movie reviews, we might enforce that the output sentiment agrees with the average of these.
To realize this intuition, we can select as final output from $\mathcal{C}_i$ the string that best aligns with this 
aggregate property (sentiment score or significance finding).
Operationally, this requires an external model to estimate
the aspect of interest as latent in a given candidate output.
This is a limitation of the approach, but in many settings it may be feasible to identify or construct a model; we were able to do so for both tasks considered here.

It may be that \emph{any} 
member of $\mathcal{C}_i$ will align well with the anticipated aggregated property. 
In such cases, we have no means of producing an output consistent with respect to synthesis, and it may be desirable to \emph{abstain} from outputting anything at all in such cases; that is, to be a \emph{cautious} summarizer \citep{ferri2004delegating,hechtlinger2018cautious}. 
We consider this strategy in the case of generating narrative synopses of evidence, as this constitutes a case in which (a) one would very much prefer not to produce a misleading summary of clinical evidence \citep{kell-etal-2021-take}, and, (b) we observe many cases where the diverse decoding strategy yields an output that seems to communicate (at a granular level) the aggregate findings expected.

\textbf{Movie Reviews} We use BERT \citep{devlin-etal-2019-bert}, fine-tuned on IMDB \citep{maas-etal-2011-learning}\footnote{\url{https://huggingface.co/lvwerra/bert-imdb}\
} to predict the sentiment inputs $x_{ij}$, using the proportion of $x_{ij} \in X_i$ with a positive score to approximate the target sentiment $z_{i\hat{y}}$. 
For each diverse prediction $\mathcal{C}_i$, we predict its sentiment $\tilde{z}_{i\hat{y}}$ via our regression model (\ref{sec:movies}), and select the prediction closest to the estimated target sentiment $|\tilde{z}_{i\hat{y}} - z_{i\hat{y}}|$. 
We find this improves model synthesis performance (Table \ref{tab:sentiment_diversity_extracted_truth}; Figure \ref{fig:improved-decoding-deltas}). 
Two authors blindly annotated 100 paired 
instances over PRIMERA generations for sentiment preference (matching the reference) between standard and diverse outputs.\footnote{Summaries were ordered by difference in extracted sentiments between base outputs and diverse outputs, then 100 instances randomly selected from the top 20\textsuperscript{th} percentile.} We find a moderate agreement Cohen's $\kappa$=0.59, and a statistically significant preference for the diverse summaries (p=0.003). %

\begin{table*}
\footnotesize
\centering
\begin{tabular}{ p{5.5em} llllll||llllll}\hline
 {}& \multicolumn{6}{c||}{Approximate Selection} & \multicolumn{6}{c}{Oracle Selection} \\
        {} & R$^2$ & $\Delta$ & PCC & $\Delta$ & R1 & $\Delta$ & R$^2$ & $\Delta$ & PCC & $\Delta$ & R1 & $\Delta$ \\ 
        \hline
        AceSum & 0.534 & 0.376 & 0.740 & 0.301 & 0.177 & \phantom{-}0.001 & 0.509 & 0.351 & 0.715 & 0.276 & 0.177 & \phantom{-}0.001 \\ 
        REFLECT$^{\text{MLE}}$ & 0.555 & 0.125 & 0.750 & 0.093 & 0.248 & \phantom{-}0.007 & 0.603 & 0.173 & 0.780 & 0.123 & 0.247 & \phantom{-}0.006 \\ 
        REFLECT$^{\text{RL}}$ & 0.406 & 0.181 & 0.638 & 0.131 & 0.222 & \phantom{-}0.004 & 0.454 & 0.229 & 0.675 & 0.168 & 0.221 & \phantom{-}0.003 \\ 
        PEGASUS & 0.649 & 0.119 & 0.809 & 0.079 & 0.248 & \phantom{-}0.003 & 0.705 & 0.175 & 0.840 & 0.110 & 0.247 & \phantom{-}0.002 \\ 
        LED & 0.653 & 0.102 & 0.815 & 0.073 & 0.241 & -0.001 & 0.711 & 0.160 & 0.847 & 0.105 & 0.240 & -0.002 \\ 
        PRIMERA & 0.685 & 0.077 & 0.833 & 0.053 & 0.254 & \phantom{-}0.000 & 0.731 & 0.123 & 0.857 & 0.077 & 0.255 & \phantom{-}0.001 \\ 
        T5-Small & 0.612 & 0.171 & 0.785 & 0.116 & 0.236 & \phantom{-}0.002 & 0.668 & 0.227 & 0.818 & 0.149 & 0.236 & \phantom{-}0.002 \\ 
        T5-Base & 0.615 & 0.099 & 0.786 & 0.066 & 0.252 & -0.001 & 0.669 & 0.153 & 0.819 & 0.099 & 0.253 & \phantom{-}0.000 \\ 
        Flan-T5-S & 0.539 & 0.127 & 0.735 & 0.088 & 0.236 & -0.001 & 0.579 & 0.167 & 0.803 & 0.156 & 0.251 & \phantom{-}0.014 \\ 
        Flan-T5-B & 0.694 & 0.097 & 0.834 & 0.060 & 0.248 & \phantom{-}0.001 & 0.741 & 0.144 & 0.861 & 0.087 & 0.248 & \phantom{-}0.001 \\ 
        Flan-T5-L & 0.732 & 0.248 & 0.866 & 0.170 & 0.243 & -0.005 & 0.875 & 0.391 & 0.937 & 0.241 & 0.244 & -0.004 \\
        Flan-T5-XL & 0.769 & 0.158 & 0.888 & 0.105 & 0.250 & -0.012 & 0.900 & 0.289 & 0.950 & 0.167 & 0.248 & -0.014 \\ 
        Reference & 0.697 & {} & 0.836 & {} & {} & {} & 0.697 & {} & 0.836 & {} & {} & {} \\ 
        \hline
    \end{tabular}

\caption{\normalsize\textbf{Movie Reviews}: Generate movie meta-reviews using standard beam search, then select using approximate (left) or oracle (right) target sentiments.
}
\label{tab:sentiment_standard_beam_search_selection}
\end{table*}

\textbf{Systematic Reviews}. For systematic reviews, we have a binary measure of \emph{significant effect} (or not).
As a proxy for $z_{i\hat{y}}$, we use {\tt RobotReviewer} to extract an effect for each of the model inputs $x_{ij}$, using the majority vote (i.e., do the plurality of $x_{ij} \in X_i$ indicate that there was an effect).
We classify each output candidate in $\mathcal{C}_i$ again using {\tt RobotReviewer} to estimate  $\tilde{z}_{i\hat{y}}$.
We then select for output the highest probability candidate in $\mathcal{C}_i$ which agrees with the majority vote of the inputs, and abstain where there are no viable candidates.
When we are able to choose a summary, we find performance similar to our measure (Table \ref{tab:cochrane_modified_generations_macro}). 

\textbf{Result}. Movie reviews show a wide range of sentiments; systematic reviews show some improvement but are biased towards no effect. %
Both settings show improvement from the switch to diverse decoding over standard beam-search methods: We repeat the generate-multiple-then-select approach with movie reviews (Table \ref{tab:sentiment_standard_beam_search_selection}) and systematic reviews (Table \ref{tab:cochrane_modified_generations_macro_standard_beam_search}). While the standard beam search did produce better overall scores when considering multiple candidates, the diverse generations produced higher correlations with human sentiment, and improved overall classification and abstention behaviors.
Both settings have some decay in overall (crude) measures of review quality - Tables \ref{tab:sentiment_diversity_extracted_truth}, \ref{tab:sentiment_standard_beam_search_selection} show small decreases in ROUGE-1 score; furthermore the diverse beam search results produce overall higher quality results (R$^2$, PCC), but how larger changes in ROUGE1 compared to a standard beam search method. Systematic Reviews behave similarly (Tables \ref{tab:cochrane_modified_generations_macro}, \ref{tab:cochrane_modified_generations_macro_standard_beam_search}), with an increase in F1 (or accuracy) comes with higher variability in ROUGE1 scores and a substantial amounts of abstention.

\begin{figure*}[t]
\centering
    \begin{subfigure}{.48\textwidth}
        \centering
        \includegraphics[width=\textwidth]{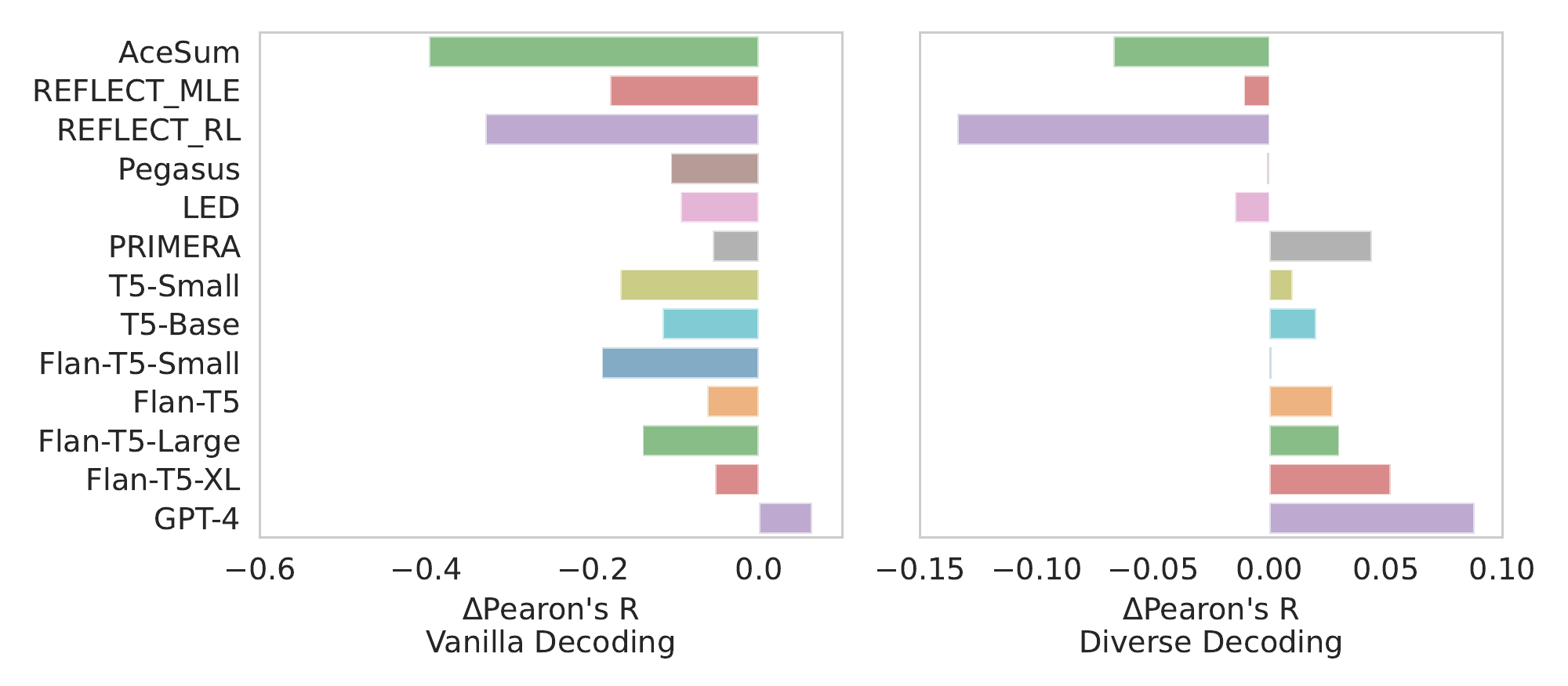}
    \end{subfigure}
    \rulesep
    \begin{subfigure}{.48\textwidth}
        \centering
        \includegraphics[width=\textwidth]{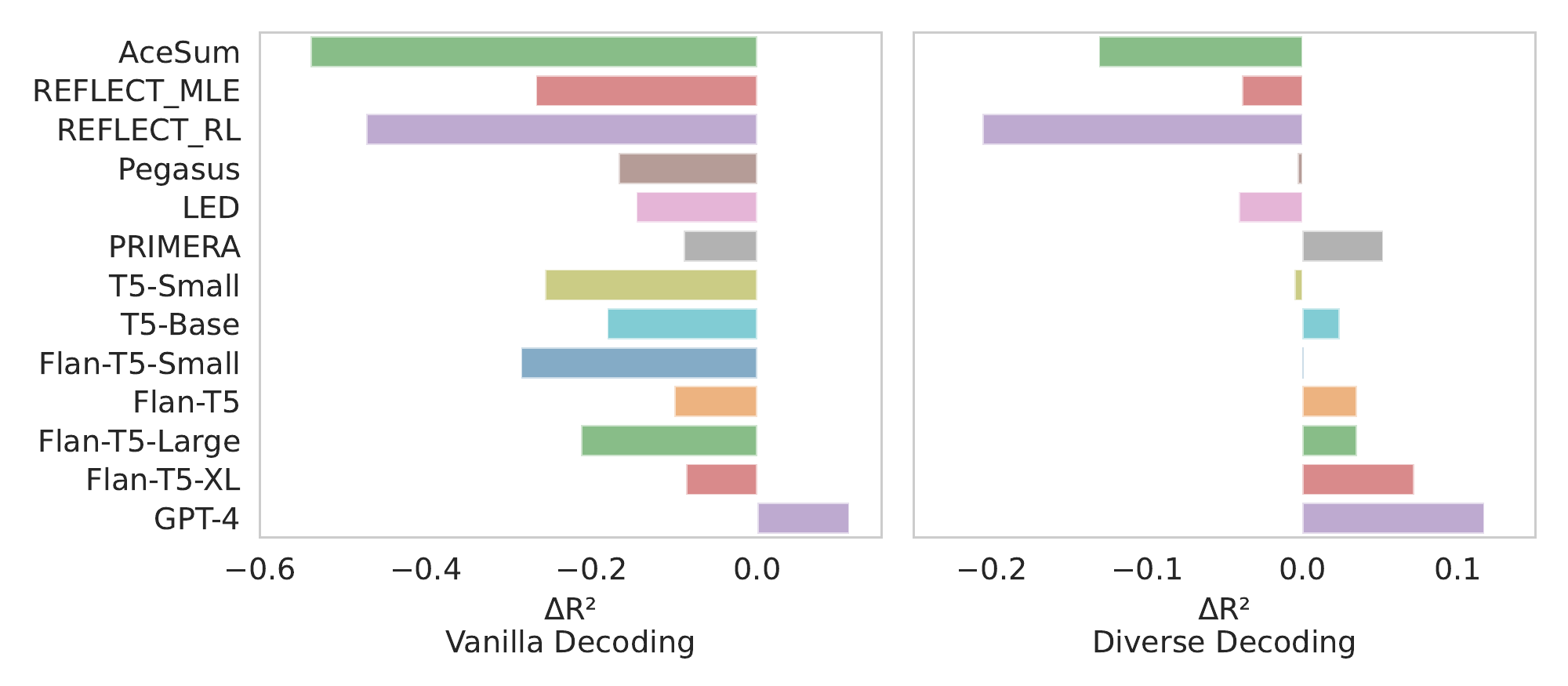}
    \end{subfigure} \vspace{-0.5em}
    \caption{Differences relative to human summaries under vanilla decoding and the proposed generate-diverse then select strategy on movie meta-reviews. We report Pearson's r (PCC) and $R^2$ as measures of synthesis ``calibration''. Vanilla decoding yields synthesis performance worse than humans, but explicitly considering synthesis at inference time results in performance comparable to and sometimes better than the human summaries (as best we can measure).}
    \label{fig:improved-decoding-deltas}
    \vspace{-.25em}
\end{figure*}

\begin{figure*}
    \centering
    \begin{subfigure}{.48\textwidth}
        \centering
        \includegraphics[width=\textwidth]{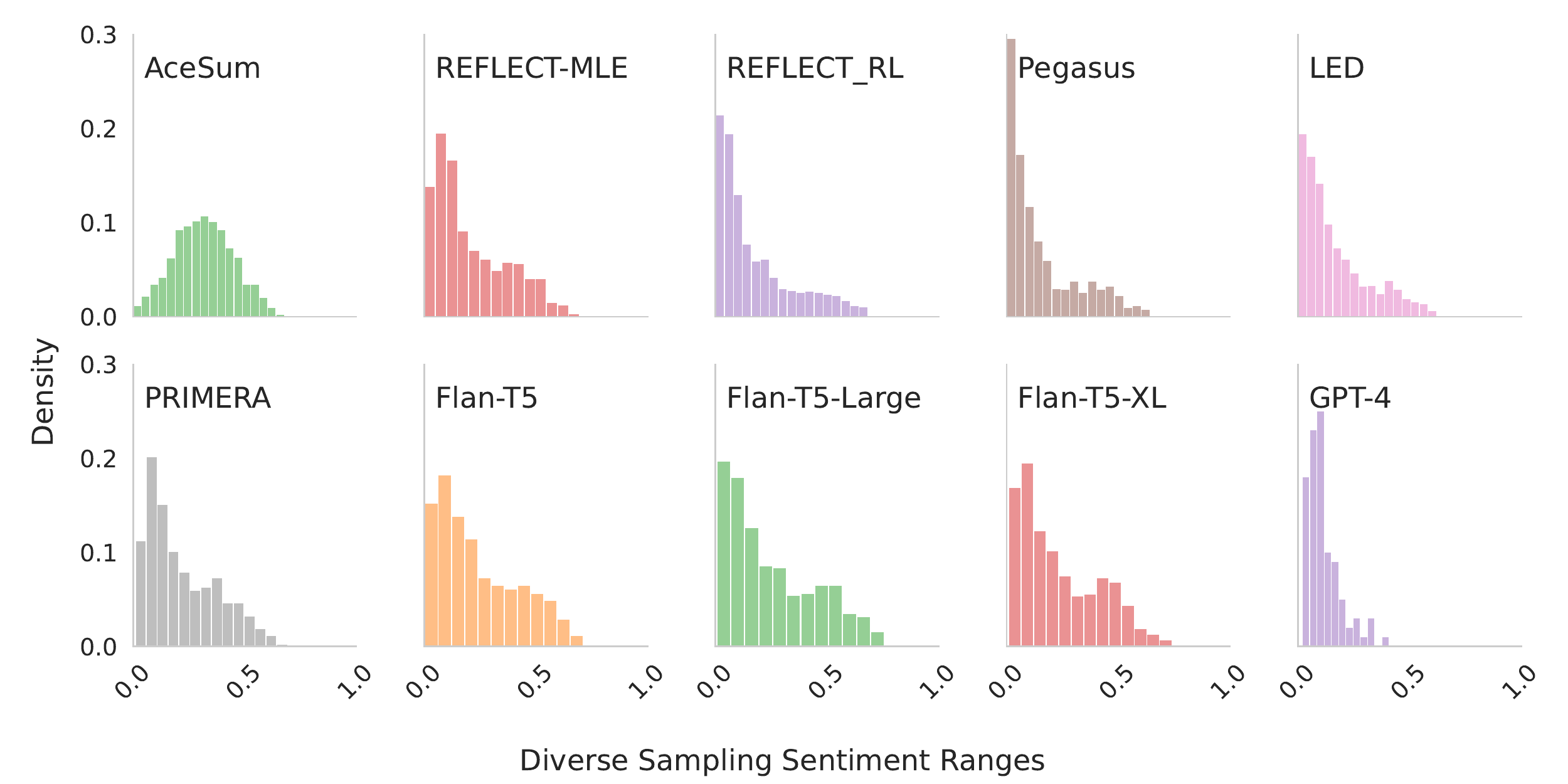}
    \end{subfigure}
    \begin{subfigure}{.48\textwidth}
        \centering
        \includegraphics[width=\textwidth]{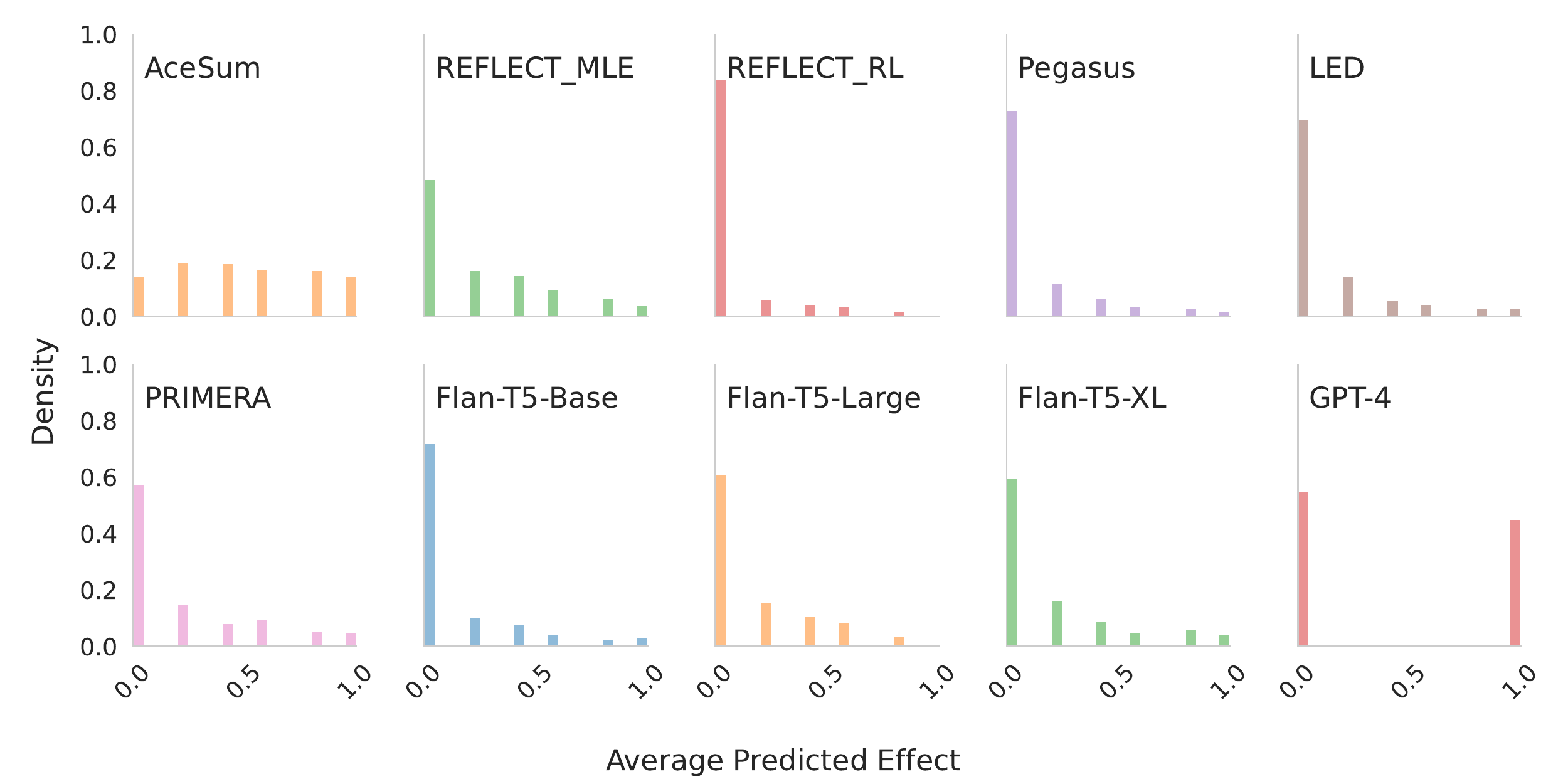}
    \end{subfigure}
    \caption{Distributions of outputs for the candiate summaries. {\bf Movie reviews} (left) show a histogram for the range of differences between lowest and highest output sentiments. {\bf Systematic reviews} (right) show histograms of the fractions of outputs reporting \emph{significant} results.}%
    \label{fig:improved_decoding_all_candidates_ranges}
    \vspace{-0.7em}
\end{figure*}

\section{Related Work} \label{sec:related_work}

\begin{table*}
    \centering
    \footnotesize
    \begin{tabular}{llllllll||lll}\hline
             & \multicolumn{7}{c||}{Multiple-then-select} & \multicolumn{3}{c}{Oracle} \\
        {} & F1 & $\Delta$ & Acc & $\Delta$ & Abs & R1 & $\Delta$ & Abs & R1 & $\Delta$ \\ \hline
        AceSum & 0.562 & 0.030 & 0.573 & \phantom{-}0.023 & 0.088 & 0.154 & \phantom{-}0.003 & 0.133 & 0.152 & \phantom{-}0.001 \\ 
        REFLECT$^{\text{MLE}}$ & 0.588 & 0.056 & 0.626 & -0.013 & 0.227 & 0.280 & \phantom{-}0.009 & 0.150 & 0.278 & \phantom{-}0.007 \\ 
        REFLECT$^{\text{RL}}$ & 0.605 & 0.100 & 0.700 & \phantom{-}0.017 & 0.430 & 0.197 & \phantom{-}0.002 & 0.247 & 0.207 & \phantom{-}0.008 \\ 
        Pegasus & 0.633 & 0.065 & 0.676 & -0.038 & 0.355 & 0.216 & \phantom{-}0.004 & 0.216 & 0.220 & \phantom{-}0.008 \\ 
        LED & 0.625 & 0.135 & 0.698 & \phantom{-}0.067 & 0.355 & 0.250 & -0.009 & 0.211 & 0.257 & -0.002 \\ 
        PRIMERA & 0.617 & 0.091 & 0.663 & \phantom{-}0.019 & 0.283 & 0.251 & -0.002 & 0.180 & 0.250 & -0.003 \\ 
        T5-Small & 0.592 & 0.052 & 0.627 & \phantom{-}0.027 & 0.211 & 0.193 & -0.012 & 0.169 & 0.190 & -0.015 \\ 
        T5-Base & 0.608 & 0.087 & 0.671 & \phantom{-}0.043 & 0.325 & 0.202 & -0.004 & 0.197 & 0.210 & \phantom{-}0.004 \\ 
        Flan-T5-S & 0.579 & 0.031 & 0.597 & \phantom{-}0.014 & 0.138 & 0.198 & \phantom{-}0.117 & 0.119 & 0.205 & \phantom{-}0.124 \\ 
        Flan-T5-B & 0.660 & 0.122 & 0.723 & \phantom{-}0.040 & 0.358 & 0.222 & \phantom{-}0.164 & 0.177 & 0.222 & \phantom{-}0.028 \\ 
        Flan-T5-L & 0.610 & 0.054 & 0.663 & -0.029 & 0.212 & 0.212 & \phantom{-}0.065 & 0.152 & 0.206 & -0.012 \\ 
        Flan-T5-XL & 0.618 & 0.131 & 0.667 & \phantom{-}0.059 & 0.300 & 0.273 & \phantom{-}0.005 & 0.189 & 0.275 & \phantom{-}0.007 \\ 
        GPT-4 & 0.653 & 0.025 & 0.640 & \phantom{-}0.000 & 0.450 & 0.275 & \phantom{-}0.002 & 0.410 & 0.269 & -0.004 \\ 
         Reference  &  {0.577}  &          & 0.686 & \\
         \hline
    \end{tabular}
            \caption{\textbf{Systematic Review} results with multiple-then-selected predictions. We report macro-averaged F1 on the set of returned results. We abstain (Abs) when no output matches the expected synthesis result. 
            }
    \label{tab:cochrane_modified_generations_macro}

\end{table*}

\textbf{Automatic (multi-document) summarization} \citep{nenkova2011automatic,maybury1999advances} has been an active subfield within NLP for decades. 
We have focused our analysis on modern, neural abstractive models for conditional text generation \citep{bahdanau2014neural}. 
In light of their empirical success, we have specifically evaluated a set of Transformer-based \citep{vaswani2017attention} models which have recently been used for multi-document summarization \citep{Beltagy2020LongformerTL,Zhang2020PEGASUSPW,xiao-etal-2022-primera,Raffel2020ExploringTL}. There has been some work on highlighting conflicting evidence in health literature specifically \citep{shah2021nutri,shah-etal-2021-nutri}, though this focused primarily on highlighting conflicting evidence and explicitly aggregating extracted content.

Multiple works have attempted gauge the difficulty of multi-document summarization. 
\citet{howmultiismds} measures the difficulty of abstractive multi-document news summarization as a function of inputs necessary to produce a final summary; they find that two to four well-chosen documents can cover a news topic sufficiently for the summarizer. 
They also find systematic reviews are particularly ill-suited to this minimal covering approach.
\citet{Giorgi2022ExploringTC} studies the impact of document retrieval behaviors on multi-document summarization performance, and find that models are sensitive to missing  inputs.

\begin{table*}
    \centering
        \footnotesize
    \begin{tabular}{llllllll||lll}\hline
                 & \multicolumn{7}{c||}{Multiple-then-select} & \multicolumn{3}{c}{Oracle} \\
        {} & F1 & $\Delta$ & Acc & $\Delta$ & Abs & R1 & $\Delta$ & Abs & R1 & $\Delta$ \\ \hline
        AceSum & 0.578 & 0.046 & 0.588 & 0.038 & 0.197 & 0.157 & \phantom{-}0.006 & 0.255 & 0.153 & -0.002 \\ 
        REFLECT$^{\text{MLE}}$ & 0.631 & 0.099 & 0.706 & 0.067 & 0.480 & 0.273 & \phantom{-}0.002 & 0.355 & 0.277 & \phantom{-}0.006 \\ 
        REFLECT$^{\text{RL}}$ & 0.603 & 0.098 & 0.753 & 0.070 & 0.483 & 0.188 & -0.011 & 0.294 & 0.201 & \phantom{-}0.002 \\ 
        Pegasus & 0.688 & 0.120 & 0.774 & 0.060 & 0.447 & 0.208 & -0.004 & 0.258 & 0.216 & \phantom{-}0.004 \\ 
        LED & 0.582 & 0.092 & 0.730 & 0.099 & 0.505 & 0.260 & \phantom{-}0.001 & 0.341 & 0.261 & \phantom{-}0.002 \\ 
        PRIMERA & 0.625 & 0.099 & 0.704 & 0.060 & 0.436 & 0.259 & \phantom{-}0.006 & 0.313 & 0.250 & -0.003 \\ 
        T5-Small & 0.603 & 0.063 & 0.633 & 0.033 & 0.258 & 0.204 & -0.001 & 0.233 & 0.201 & -0.004 \\ 
        T5-Base & 0.613 & 0.092 & 0.692 & 0.064 & 0.405 & 0.208 & \phantom{-}0.002 & 0.300 & 0.211 & \phantom{-}0.005 \\ 
        Flan-T5-S & 0.603 & 0.055 & 0.632 & 0.049 & 0.361 & 0.081 & \phantom{-}0.000 & 0.333 & 0.080 & -0.001 \\ 
        Flan-T5-B & 0.637 & 0.099 & 0.761 & 0.078 & 0.500 & 0.195 & \phantom{-}0.001 & 0.300 & 0.198 & \phantom{-}0.004 \\ 
        Flan-T5-L & 0.673 & 0.117 & 0.771 & 0.079 & 0.478 & 0.177 & -0.041 & 0.281 & 0.174 & -0.044 \\ 
        Flan-T5-XL & 0.594 & 0.107 & 0.665 & 0.057 & 0.394 & 0.271 & \phantom{-}0.003 & 0.311 & 0.269 & \phantom{-}0.001 \\ 
         Reference  &  {0.577}  &          & 0.686 & \\
        \hline
    \end{tabular}
        \caption{\normalsize\textbf{Systematic reviews} results with multiple generate-then-select predictions, this time using the top-5 results from standard beam-search.
    } %
    \label{tab:cochrane_modified_generations_macro_standard_beam_search}
\end{table*}

\begin{table}
    \small
    \begin{tabular}{ll}\hline
Summary &  Sent. \\\hline
  \parbox[t]{.375\textwidth}{You Don't Mess With the Zohan's handful of laughs are almost enough to compensate for its inconsistent tone and stale, obvious jokes.} &  0.243 \\
\parbox[t]{.375\textwidth}{You Don't Mess with the Zohan has a handful of crotch thrusts, but not enough of them land.} &  0.429 \\
\parbox[t]{.375\textwidth}{You Don't Mess With the Zohan's handful of laughs are almost enough to compensate for its aimless, crass script.}&  0.288 \\
          \parbox[t]{.375\textwidth}{You Don't Mess with the Zohan has its moments, but not all of them -- and the jokes are embarrassingly crass and often crude.} &  0.434 \\
\parbox[t]{.375\textwidth}{\textbf{You Don't Mess with the Zohan has its moments, but not all of them -- and the jokes are embarrassingly crass and often crude. The script}} &  0.406 \\\hline
\end{tabular}
    \caption{Diverse meta-review generations and automatically inferred sentiment scores for ``You Don't Mess With The Zohan''. Target meta-review sentiment of 37\%: We {\bf bold} the closest generation in terms of (inferred) sentiment.}
    \label{tab:diverse_movies_example2}
\end{table}

\textbf{Sentence fusion} One view on synthesis might be that is a particular kind of \emph{sentence fusion} \citep{barzilay-mckeown-2005-sentence}. 
However, past work on ``fusing'' sentences has assumed that the aim is to generate an output that contains the information common to similar sentences \citep{thadani2013supervised}.
This is intuitive in the context of, e.g., summarizing multiple news articles covering the same event. 
But here we are interested in the more challenging setting in which the output should reflect an aggregate measure of potentially conflicting evidence or opinions.

\textbf{Review and opinion summarization} considers a similar task to ours: Aggregating (usually product) reviews and opinions into a single coherent text. \citet{oved-levy-2021-pass} developed a system with a similar generate-then-select approach, however this work was focused on generating \emph{plausible summaries} rather than accurate \emph{syntheses}, by selecting amongst candidates via a voting mechanism designed to mimic human preferences. Other related work has considered generating personalized and/or aspect-oriented summaries \cite{he-etal-2017-unsupervised,angelidis-lapata-2018-summarizing,amplayo-lapata-2020-unsupervised,amplayo-lapata-2021-informative,amplayo-etal-2021-aspect,angelidis-etal-2021-extractive}. \citet{amplayo-lapata-2021-informative} propose a T5 variant for pooling instance representations, and also use Rotten Tomatoes as a dataset. 
This work (and \citealp{amplayo-etal-2021-aspect}) includes a manual evaluation of how well system summaries are \textit{supported} by input reviews, in contrast to how well a summary agrees with \textit{all} inputs in the precise sense we have considered. 
We note that none of these prior works directly \textit{probe} model responsiveness to changes in input composition.

Also related is the work of \citet{pmlr-v97-chu19b}, which considered \emph{unsupervised} approaches to multi-document summarization of Yelp! and Amazon reviews; they adopt an auto-encoder that  ``decodes'' the mean of input representations to target summaries. They similarly note that output texts should convey mean input sentiment, and report ``sentiment accuracy'' as one of their metrics. 
But the synthesis aspect is not their main focus, and they consider only unsupervised settings (rather than the SOTA fine-tuned summarization models we have evaluated).

\textbf{Interpretation and analysis of neural models for NLP} This work is also related to the emerging body of work on analyzing neural NLP models, their behaviors, ``knowledge'', and ``abilities'' in general, e.g., \cite{linzen-etal-2016-assessing,tenney2018what,petroni-etal-2019-language,niven-kao-2019-probing,meng2022locating}. 
There has been some work specifically on analyzing neural summarization models. \citet{xu-etal-2020-understanding-neural} investigated when a model is likely to copy rather than generate. 
\citet{xu-durrett-2021-dissecting} assessed when models were relying on the local input to produce particular output tokens, and when they instead rely mostly on a background language distribution acquired in pre-training. In contrast to \citet{Giorgi2022ExploringTC} we explore beyond surface forms and explore the specific aspect of text \textit{synthesis}.

\textbf{Factuality of neural summarizers} Neural conditional generation models have proven adept at producing fluent outputs, but when summarizing they are prone to \emph{hallucinating} content unsupported by input documents \citep{maynez-etal-2020-faithfulness,kryscinski-etal-2019-neural}. 
Automated metrics such as ROUGE do not reliably capture such phenomena \citep{falke-etal-2019-ranking,maynez-etal-2020-faithfulness}. 
This has motivated the design of automated factuality metrics, e.g., \cite{wang-etal-2020-asking,xu-etal-2020-fact}; see \citet{pagnoni-etal-2021-understanding} for an overview.

\section{Conclusions} \label{sec:conclusionS}

We have outlined and investigated the problem of \emph{synthesis} as related to some summarization tasks.
We showed that existing models are partially able to synthesize implicitly, but do so imperfectly: 
the aggregation they perform is sensitive to input ordering, and they are not as sensitive to perturbations in the composition of inputs as one would hope. 
Some models specifically designed for these tasks (AceSum, QT, REFLECT) are \textit{less} sensitive to these perturbations, but offer worse overall performance than an equivalently sized transformer model (compare LED and REFLECT - REFLECT integrates a model with the same base LLM parameters as a portion of its synthesis model).
Furthermore, increasing model size within an architecture can lead to fairly substantial improvements (LED to PRIMERA, T5 Small to Base, similarly for Flan-T5). Pretraining methods have some impact as well: T5 and Flan-T5 do not perform identically despite an identical model structure and comparable sizes, and GPT-4 clearly outperforms all models in this case, including the bespoke ones.

We proposed and validated a straightforward inference time method to improve model synthesis capabilities by preferentially outputting summary candidates that align with a predicted aggregate measure, and demonstrated empirically that this offers gains in performance. 
These gains are primarily limited by the underlying models' behaviors, but potentially bring performance on these single, task-specific metrics, on par to human performance, when the model is capable of providing a response that aligns with the proxy metrics.

We hope this work encourages additional research into summarization models that explicitly optimize to accurately synthesize potentially conflicting evidence. %
We are particularly interested in understanding \textit{why} models fail to synthesize --- they clearly learn to produce synthesis-like text, but fail to yield the best option, even among their top candidates. We use summary reranking as a means to surface these more-appropriate summaries, but this is solely post-hoc as opposed to controlling for a more suitable generation, or ideally improving base model performance.

Our methods focus solely on improving performance at single specific task measures, potentially at a cost to other review qualities. Users of such systems may have auxiliary goals, perhaps requiring multiple measures of synthesis quality, other measures of overall review quality, or a greater (or lesser) willingness to abstain. Abstinence can be a feature beyond the case of systematic reviews; systems may have other specific rules for when to abstain: e.g. toxic language, challenging to verify statements, or distance from an overall objective (i.e. abstaining in the movie reviews case).

This work has several limitations. We have made an effort to fine-tune several popular summarization models, but limited our analysis to models of relatively modest size (due to the GPU memory required to train long sequence summarization models). These behaviors appear to change with larger models (e.g. the small vs base-sized models, GPT-4 \cite{openai2023gpt4}), but building robustness to perturbations while maintaining sensitivity to input composition is a non-obvious challenge.
We also have reported results on only English-language tasks. %
Finally, we focused on a relatively narrow behavior (synthesis of a single aspect); models may succeed in this respect while failing in other ways.

\clearpage

\section*{Acknowledgements}
This work was supported by the National Science Foundation (NSF) RI 2211954, and by the National Institutes of Health (NIH) under the National Library of Medicine (NLM), R01-LM012086-01A1. We thank the Northeastern University Research Computing Discovery Cluster. We thank the anonymous reviewers for their feedback.

\bibliography{anthology,custom}
\bibliographystyle{acl_natbib}

\clearpage

\appendix

\begin{table*}[H]
    \begin{tabular}{ll}\hline
Generated &       Effect \\\hline
                                 \parbox[t]{.7\textwidth}{  Ketanserin versus placebo in the Raynaud's phenomenon is neither effective nor safe. The Raynaud's phenomenon is associated with significant adverse effects including dizziness and pain. The effectiveness of ketanserin for the Raynaud's phenomenon is unknown. } & no significant difference \\
                                                                                                            \parbox[t]{.7\textwidth}{Ketanserin versus placebo in the Raynaud's phenomenon is neither effective nor safe. The Raynaud's phenomenon is associated with significant adverse effects including dizziness and pain.} & no significant difference \\
                                                                                                \parbox[t]{.7\textwidth}{  Ketanserin and serotonin receptor antagonists in the Raynaud's phenomenon treatment of systemic scleroderma reduce the incidence of ischaemic ulcers and may reduce the frequency of adverse events.} &   significant difference \\
\parbox[t]{.7\textwidth}{The Raynaud's phenomenon is associated with a small number of adverse effects when administered orally to patients with Raynaud's phenomenon. The frequency of Raynaud's phenomenon is similar to that of other drugs. However, there is little evidence to aid the treatment of Raynaud's phenomenon.} & no significant difference \\
                                                                               \parbox[t]{.7\textwidth}{ The Raynaud's phenomenon is associated with a small number of adverse effects when administered orally to patients with Raynaud's phenomenon. The frequency of Raynaud's phenomenon is similar to that of other drugs. } & no significant difference \\\hline
\end{tabular}

    \caption{An instance where generating multiple reviews allows our models to find a candidate summary reporting a significant difference (the target).
    }
    \label{tab:diverse_cochrane_generated}
\end{table*}

\begin{table*}[H]
    \centering
    \begin{tabular}{ll}\hline
 Generated &       Effect \\\hline

   \parbox[t]{.7\textwidth}{The overall evidence supports the use of topical antibiotics in surgical patients who have undergone minor surgery, compared to no treatment. The effect on other outcomes, other than infection rate, is consistent. The safety profile of topical antibiotics is also of concern. Further well-designed RCTs are needed to assess effectiveness of topical antibiotics in surgical patients.} & no significant difference \\
                                             \parbox[t]{.7\textwidth}{                                        A single application of topical antibiotics in surgical site wounds reduces the risk of infection, and the risk of other complications, including wound dehiscence. The risk of infection recurrence is low. The use of topical antibiotics outside of surgery should be restricted to surgical site wounds.} & no significant difference \\
                                                                                                                                                                               \parbox[t]{.7\textwidth}{ A single application of topical antibiotics in surgical site wounds reduces the risk of infection, and the risk of other complications, including wound dehiscence. The risk of infection recurrence is low. }& no significant difference \\
                                                                                            \parbox[t]{.7\textwidth}{The overall evidence supports the use of topical antibiotics in surgical patients to reduce the risk of infection, and the risk of other complications, especially in high-risk patients. There is a lack of evidence in low-risk patients to support the use of topical antibiotics in this setting. } &   significant difference \\
\parbox[t]{.7\textwidth}{A single application of topical antibiotics in surgical site infection prevention has been demonstrated to reduce the risk of infection in patients who have undergone surgery. The number of patients who have been treated with topical antibiotics has been small but this is due to risk of bias in the trials. Ointment use should be limited to patients whose primary wound is irradiated.} &   significantly difference \\\hline
\end{tabular}
\caption{An instance where generating multiple reviews allows our models to find a candidate summary reporting a significant difference (the target).}
    \label{tab:my_label}
\end{table*}

\end{document}